\documentclass[lettersize,journal]{IEEEtran}
\usepackage{amsmath,amsfonts}
\usepackage{algorithmic}
\usepackage{algorithm}
\usepackage{array}
\usepackage[caption=false,font=normalsize,labelfont=sf,textfont=sf]{subfig}
\usepackage{textcomp}
\usepackage{stfloats}
\usepackage{url}
\usepackage{cite}
\usepackage{verbatim}

\usepackage[colorlinks,
            linkcolor=blue,
            anchorcolor=blue,
            citecolor=blue]{hyperref}

\usepackage{graphicx}
\usepackage{amssymb}
\usepackage{booktabs}
\usepackage{array}
\usepackage{color}
\usepackage{colortbl} 
\usepackage{svg}
\usepackage{float}
\usepackage{subcaption}
\usepackage{multirow}
\usepackage{makecell}
\usepackage{bm}
\usepackage{float}
\usepackage[usestackEOL]{stackengine}
\usepackage{marvosym, ifsym} 

\newcommand{\smallTitle}[1]{{\noindent\textbf{{#1}}}}

\newcommand{\eg}{\emph{e.g.}, }
\newcommand{\ie}{ \emph{i.e.}, }

\begin{document}

\title{VBench-2.0: Advancing Video Generation Benchmark Suite for Intrinsic Faithfulness}

\author{Dian Zheng*, Ziqi Huang*, Hongbo Liu, Kai Zou, Yinan He, Fan Zhang,\\ Lulu Gu, Yuanhan Zhang, Jingwen He, Wei-Shi Zheng~\textsuperscript{\Letter}, Yu Qiao~\textsuperscript{\Letter}, Ziwei Liu~\textsuperscript{\Letter}

\thanks{* equal contribution. \textsuperscript{\Letter} corresponding authors.}
\thanks{Email: Dian Zheng zd1423606603@gmail.com and Ziqi Huang ziqi002@ntu.edu.sg}
\thanks{D. Zheng, H. Liu, K. Zou, F. Zhang, L. Gu, J. He, Y. Qiao are with Shanghai Artificial Intelligence Laboratory. Z. Huang, Y. Zhang, Z. Liu are with the S-Lab, Nanyang Technological University. W. Zheng is with Sun Yat-sen University. J. He is also with The Chinese University of Hong Kong.}
}



\maketitle

\begin{abstract}
Video generation has advanced significantly, evolving from producing unrealistic outputs to generating videos that appear visually convincing and temporally coherent. To evaluate these video generative models, benchmarks such as VBench have been developed to assess their faithfulness, measuring factors like per-frame aesthetics, temporal consistency, and basic prompt adherence. However, these aspects mainly represent \textbf{superficial faithfulness}, which focus on whether the video appears visually convincing rather than whether it adheres to real-world principles. 
While recent models perform increasingly well on these metrics, they still struggle to generate videos that are not just visually plausible but fundamentally realistic.
To achieve real ``world models'' through video generation, the next frontier lies in \textbf{intrinsic faithfulness} to ensure that generated videos adhere to physical laws, commonsense reasoning, anatomical correctness, and compositional integrity. Achieving this level of realism is essential for applications such as AI-assisted filmmaking and simulated world modeling.

To bridge this gap, we introduce \textbf{VBench-2.0}, a next-generation benchmark designed to automatically evaluate video generative models for their \textbf{intrinsic faithfulness}. VBench-2.0 assesses five key dimensions: Human Fidelity, Controllability, Creativity, Physics, and Commonsense, each further broken down into fine-grained capabilities. Tailored to individual dimensions, our evaluation framework integrates generalists such as state-of-the-art VLMs and LLMs, and specialists, including anomaly detection methods proposed for video generation. We conduct extensive human preference annotations to ensure evaluation alignment with human judgment.
By pushing beyond superficial faithfulness toward intrinsic faithfulness, VBench-2.0 aims to set a new standard for the next generation of video generative models in pursuit of intrinsic faithfulness.
\end{abstract}

\begin{IEEEkeywords}
Video Generative Models, Evaluation Benchmark.
\end{IEEEkeywords}

\begin{figure*}[t]
  \centering
   \includegraphics[width=1.0\linewidth]{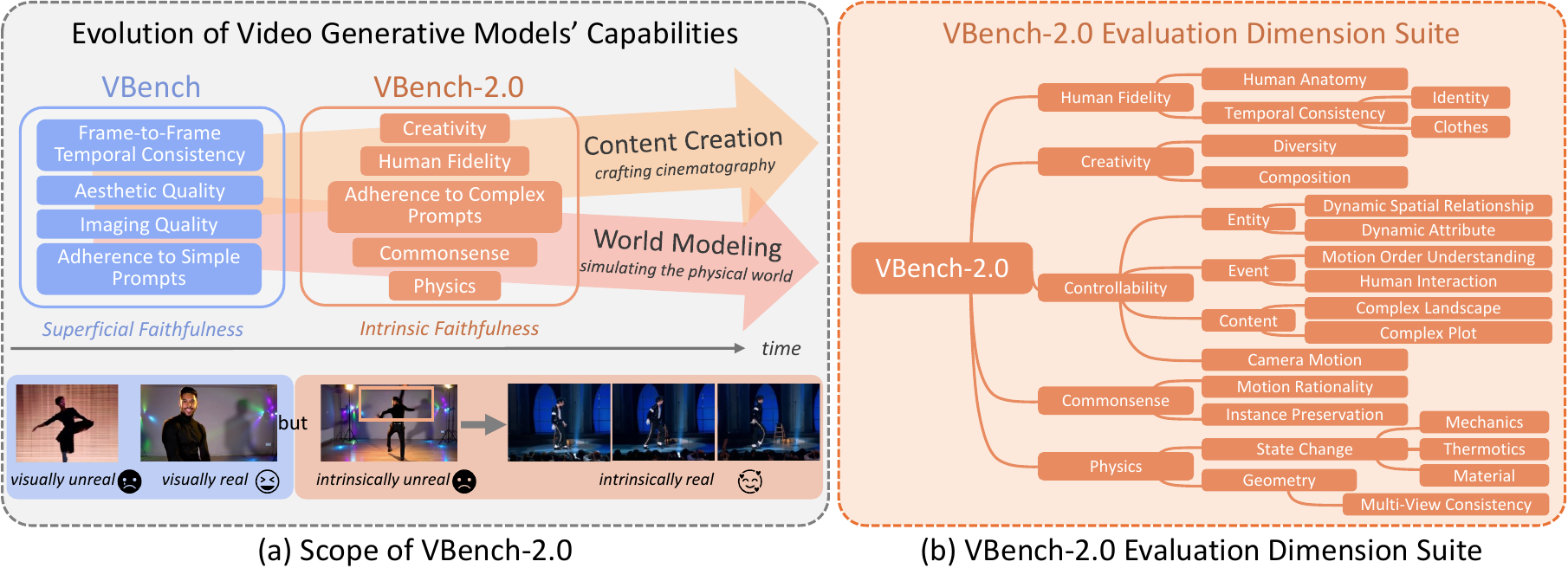}
   \caption{
   \textbf{Overview of VBench-2.0.}  
        \textbf{(a) Scope of VBench-2.0.} Video generative models have progressed from achieving \textit{superficial faithfulness} in fundamental technical aspects such as pixel fidelity and basic prompt adherence, to addressing more complex challenges associated with \textit{intrinsic faithfulness}, including commonsense reasoning, physics-based realism, human motion, and creative composition. While VBench primarily assessed early-stage technical quality, VBench-2.0 expands the benchmarking framework to evaluate these advanced capabilities, ensuring a more comprehensive assessment of next-generation models.  
        \textbf{(b) Evaluation Dimension of VBench-2.0.} 
        VBench-2.0 introduces a structured evaluation suite comprising five broad categories and 18 fine-grained capability dimensions. 
   }
    \label{fig:teaser}
\end{figure*}

\section{Introduction}
\IEEEPARstart{V}{ideo} generation aims to create realistic and temporally coherent video sequences, with a wide range of applications in video editing~\cite{liew2023magicedit,chai2023stablevideo,tokenflow2023,huang2023inve,couairon2023videdit,liu2023videop2p,zhang2023consistent,zhao2023controlvideo,vid2vid-zero,ceylan2023pix2video,qi2023fatezero,lee2023textvideoedit,zhao2023makeaprotagonist,yang2025videograin}, customization~\cite{kumari2022customdiffusion,dreampose_2023,he2023animate}, image animation~\cite{guo2023animatediff, xing2023dynamicrafter}, and world models~\cite{agarwal2025cosmos}.

Earlier video generative models~\cite{wang2023lavie, wang2023modelscope, chen2023videocrafter1, hong2022cogvideo} primarily focused on generating short video clips of around two seconds, emphasizing fundamental capabilities like per-frame aesthetics and temporal consistency. To systematically evaluate these capabilities, benchmarks~\cite{huang2024vbench, huang2024vbench++, liu2023evalcrafter} such as VBench~\cite{huang2024vbench, huang2024vbench++} have been developed to assess aspects like per-frame aesthetics, frame-to-frame temporal smoothness, and  adherence to simple text prompts, which we refer to as \textbf{\textit{superficial faithfulness}}, the degree to which generated videos appear visually convincing. As video generative models continue to evolve, recent state-of-the-art models, including Sora~\cite{sora}, Kling~\cite{kling}, Gen-3~\cite{Gen3}, HunyuanVideo~\cite{kong2024hunyuanvideo}, and Veo 2~\cite{Veo2}, have demonstrated strong performance on these metrics, and many aspects of superficial faithfulness are now approaching saturation.

However, as video generation moves towards more advanced applications, particularly in areas that require AI models to simulate and reason about the real world~\cite{agarwal2025cosmos}, such as AI-driven storytelling and video-generation-based simulation, the new frontier shifts from \textit{merely appearing real} to \textit{being intrinsically real}. We term this as
\textit{\textbf{intrinsic faithfulness}} - a concept that extends beyond per-frame quality and smooth motion, requiring that generated videos adhere to deeper principles such as physical laws, commonsense reasoning, anatomical correctness, and compositional integrity. Achieving this level of faithfulness is essential for applications ranging from AI-assisted filmmaking to virtual environments for embodied intelligence, ultimately paving the way for the development of true world models that can accurately represent and predict real-world dynamics.

\begin{figure}[t]
  \centering
   \includegraphics[width=0.85\linewidth]{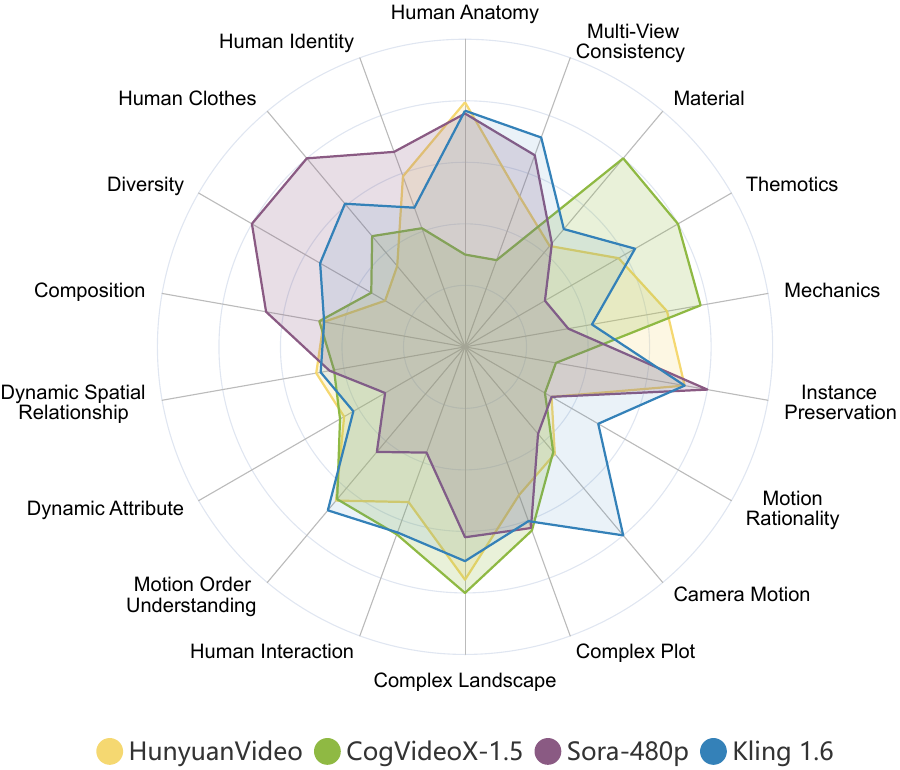}
   \caption{
   \textbf{VBench-2.0 Evaluation Results of SOTA Models.} The figure presents the evaluation results of four recent state-of-the-art video generation models across 18 VBench-2.0 dimensions. The results are normalized per dimension for a clearer comparison. For detailed numerical results, refer to Table~\ref{tab:raw_metrics}. 
   }
    \label{fig:fig_paper_radar_big}
\end{figure}

To drive video generative models towards this next generation of capabilities, we introduce \textbf{VBench-2.0}, a benchmark suite designed to evaluate video generative models along five emerging dimensions beyond superficial faithfulness: \textit{Human Fidelity}, \textit{Controllability}, \textit{Creativity}, \textit{Physics}, and \textit{Commonsense}. Each dimension is further broken down into isolated sub-abilities (shown in Figure~\ref{fig:teaser}(b)), providing a \textit{fine-grained assessment} of the \textbf{\textit{intrinsic faithfulness}} of video generative models. Given the complexity of these evaluations, we leverage \textit{generalists} such as state-of-the-art Video Language Models (VLMs) and Large Language Models (LLMs) to perform structured reasoning and judgment. Specifically, we design two complementary evaluation methods: \textit{1)} \textit{text description alignment} to assess abstract concept and semantic understanding, leveraging modern VLM's strong captioning and LLM's reasoning ability, and \textit{2)} \textit{video-based multi-question answering} for basic visual understanding. To enhance evaluation robustness in specific domains, we incorporate \textit{specialists}, such as human anomaly detection pipelines trained for generated videos. Furthermore, we use \textit{evaluation safeguards} like \textit{pre-filtering}, and \textit{redundant questioning} to mitigate hallucinations and inconsistencies in our tailored evaluation pipelines for each dimension. Additionally, we follow VBench~\cite{huang2024vbench, huang2024vbench++} and conduct extensive human preference annotations to validate and align our automated evaluation results with human judgment.

Our evaluation provides \textit{comprehensive insights} into the strengths and weaknesses of state-of-the-art video generative models. While recent models demonstrate emerging abilities in human anatomy, consistency, and some degree of novel creativity, they still struggle with generating complex plots, handling simple dynamic changes in objects, and remain unstable in commonsense reasoning, highlighting key open challenges in video generation towards synthesizing the world with \textit{intrinsic faithfulness}. We provide an in-depth discussion on possible causes, potential solutions, inherent trade-offs, and future work in Section~\ref{sec:insights}.

VBench-2.0 will be fully open-sourced, being complementary to VBench~\cite{huang2024vbench, huang2024vbench++}, and providing a standardized framework for evaluating future breakthroughs in video generation. While VBench remains essential for assessing \textit{superficial faithfulness}, VBench-2.0 extends the evaluation scope to \textit{intrinsic faithfulness}, addressing deeper aspects of video realism. We will continually integrate newly released video generative models into VBench-2.0. By setting a higher standard for evaluation, VBench-2.0 aims to play a pivotal role in guiding the development of next-generation video generative models. Together, VBench and VBench-2.0 form a comprehensive benchmarking system, driving the field beyond superficial faithfulness towards truly intrinsically faithful video generation.


\section{Related Works}
\subsection{Video Generative Models} With the advancements in diffusion models~\cite{sohl2015deep, song2020score, ho2020ddpm, song2020ddim, zhang2023controlnet, blattmann2023stable, esser2024scaling, mou2023t2iadapter, huang2023collaborative, ding2021cogview, ding2022cogview2, imagenvideo}, variational autoencoder-based compression techniques~\cite{kingma2013vae, van2017vqvae, esser2021vqgan, podell2023sdxl, yu2023magvit}, and transformer architectures~\cite{dosovitskiy2020image, peebles2022scalable}, video generation has emerged as one of the most dynamic frontiers in artificial intelligence research. Prior to Sora's breakthrough, predominant text-to-video models primarily focused on synthesizing short video clips (2-3 seconds duration)~\cite{chai2023stablevideo, wang2023lavie, guo2023animatediff, hong2022cogvideo, luo2023videofusion, he2022lvdm, zhou2023magicvideo, zhang2023show1, ge2023pyoco, blattmann2023videoldm, wang2023modelscope, khachatryan2023text2videozero, chen2023videocrafter1, yang2024cogvideox, moviegen, sora, Veo2} through incremental improvements in visual fidelity and temporal consistency. Sora~\cite{sora} pioneered the scaling paradigm in video generation by demonstrating unprecedented model capacity through large-scale training, paving the way for the development of next-generation video foundation models~\cite{yang2024cogvideox, Gen3, kling, kong2024hunyuanvideo, wan2.1, ma2025stepvideot2vtechnicalreportpractice, agarwal2025cosmos, Minmax, fan2025vchitect, si2025RepVideo, opensora2} that achieve remarkable visual quality and robust spatiotemporal coherence. They have shifted focus toward enhancing video generation adhere to deeper principles such as physical laws and commonsense reasoning that focuses more on action continuity and realistic physical-world perception~\cite{yang2024cogvideox, sora, agarwal2025cosmos}, or high-quality human-centric generation with creativity potential~\cite{kling, kong2024hunyuanvideo}. 
Existing benchmarks cannot systematically evaluate these new explorations, and VBench-2.0 takes the initiative to provide a comprehensive benchmark for evaluating emerging capabilities towards the goal of achieving intrinsic faithfulness through video generation.

\subsection{Evaluation of Video Generative Models}
Initially, video generative models primarily relied on conventional evaluation metrics such as Fr\'echet inception distance (FID)~\cite{heusel2017fid}, Inception Score (IS)~\cite{salimans2016inceptionscore}, and Fr\'echet video distance (FVD)~\cite{unterthiner2019fvd}. However, these metrics provided limited insight into the diverse and complex capabilities of modern video generation. 
Recent evaluation frameworks~\cite{huang2024vbench, huang2024vbench++, liu2023evalcrafter, liu2023fetv, zhang2024evaluationagent} such as VBench~\cite{huang2024vbench, huang2024vbench++} introduced a more structured approach by disentangling evaluation into multiple capability dimensions, enabling more detailed and interpretable assessments.
These benchmarks focus on fundamental technical attributes such as per-frame quality, temporal consistency, and basic prompt adherence. However, as models continue to improve, certain dimensions within VBench begin to saturate, necessitating broader evaluation scope that assess deeper aspects of intrinsic faithfulness in video generation.
To address this, specialized benchmarks have emerged. PhyGenBench~\cite{phygenbench} evaluates a model’s understanding of physical laws through Vision-Language Models (VLMs). T2V-CompBench~\cite{sun2024t2v} assesses compositionality, including motion, actions, spatial relationships, and attributes. StoryEval~\cite{wang2024your} focuses on storytelling capabilities by aggregating the responses from two VLMs.
Unlike prior benchmarks, which either focus on fundamental capabilities~\cite{liu2023evalcrafter, huang2024vbench, huang2024vbench++} or specific emerging domains~\cite{phygenbench, sun2024t2v}, VBench-2.0 introduces a comprehensive framework to systematically evaluate next-generation video generation capabilities, bridging the gap in the evolving landscape of video generation.

\begin{table*}[t]
\small
\centering
\caption{\textbf{Comparison of Video Generation Benchmarks.} We compare existing video generation benchmarks based on their evaluation aspects. VBench-2.0 is the first comprehensive benchmark to assess intrinsic faithfulness in video generation, complementing VBench~\cite{huang2024vbench, huang2024vbench++}. Detailed aspects include per-frame quality (Frame Wise), temporal consistency (Temp Cons), adherence to simple prompts (Simp Pmpt), compositional creativity (Comp Crea), commonsense reasoning (Com Sense), physics-based realism (Phy), human anatomy (Human Anat), and adherence to complex prompts (Cplx Pmpt).}
\begin{tabular}{c|ccc|ccccc}
\toprule[0.15em]
\textbf{} & \multicolumn{3}{c|}{\textbf{Superficial Faithfulness}} & \multicolumn{5}{c}{\textbf{Intrinsic Faithfulness}} \\
\textbf{} & \textit{Frame Wise} & 
\textit{Temp Cons} & 
\textit{Simp Pmpt} & 
\textit{Comp Crea} &
\textit{Com Sense} &
\textit{Phy} & 
\textit{Human Anat} & 
\textit{Cplx Pmpt} \\ 
\midrule[0.12em]
VBench~\cite{huang2024vbench, huang2024vbench++}  & \checkmark & \checkmark & \checkmark &  & &  & &    \\
T2V-CompBench~\cite{sun2024t2v}   &  &  & \checkmark & \checkmark & \checkmark &  &   &  \\
PhyGenBench~\cite{phygenbench}   &  & & &  &  & \checkmark  &  &    \\
StoryEval~\cite{wang2024your}   &  & \checkmark & &  &  &  &  &  \checkmark  \\
\textbf{VBench-2.0 (Ours)}    & & \checkmark & & \checkmark & \checkmark & \checkmark  & \checkmark & \checkmark \\ \bottomrule[0.12em]
\end{tabular}
\label{tab:benchmark_comparison}
\end{table*}

\section{VBench-2.0 Suite for Intrinsic Faithfulness}

In this section, we introduce the evaluation framework of VBench-2.0. Section~\ref{subsec:evaluation_dimension_suite} presents  the five key evaluation dimensions and their respective assessment methods. Unlike VBench, which primarily evaluates \textit{superficial faithfulness}, VBench-2.0 introduces a suite of tests to assess \textit{intrinsic faithfulness}, focusing on deeper properties such as physics, commonsense, and creativity, human, and controllability. To ensure robust evaluation, we integrate multiple assessment methodologies, including LLM-assisted text alignment, video-based multi-question answering, and specialist models trained for anomaly detection. Prompt suite is introduced in Section~\ref{subsec:prompt_suite} and Section~\ref{subsec:human_preference_annotation} describes the human annotation pipeline of VBench-2.0, including the collection and processing of video data, labeling of human annotations, and the definition of scoring criteria.

\subsection{Evaluation Dimension Suite}
\label{subsec:evaluation_dimension_suite}

VBench-2.0 evaluates video generation along five key dimensions: \textit{Human Fidelity, Creativity, Controllability, Physics, and Commonsense}. Each dimension is further decomposed into sub-dimensions, ensuring a fine-grained assessment of a model’s capabilities. We employ a structured approach that combines \textit{generalist} reasoning models (VLMs/LLMs) with \textit{specialist} detectors.
When using \textit{generalist}, we adopt two evaluation schemes where each is tailored to different types of semantic understanding required across evaluation dimensions.

\smallTitle{Text Description Alignment.} This scheme is suitable for complex or subtle scenarios, such as those involving nuanced human interactions or multi-step plots. In these cases, the composite scene is broken down and interpreted step by step by the Vision-Language Model (VLM), which generates a descriptive caption guided by system prompts that focused on specific aspects of the video (\eg prompting only about human interactions or a specific part of the plots). The correctness of the generated content is then judged by a Large Language Model (LLM), which compares the VLM-generated caption with a ground-truth reference. The reference may be the original text prompt, a predefined answer, or relevant metadata. This process is formalized as: \begin{align} Answer=LLM(VLM(V|S_v), T~|~S_l), \end{align} where \textit{V} is the generated video, \textit{T} denotes the text prompt or crafted reference, and $S_v$ and $S_l$ are the system prompts for the VLM and LLM, respectively. The LLM outputs a binary judgment (“yes” or “no”) that is discretized to a score of 1 or 0 to reflect caption-reference matching. This scheme excels in semantic understanding dimensions like \textit{Complex Plot} and \textit{Human Interaction}, where VLMs usually struggle with high-level interpretation, but LLMs demonstrate stronger reasoning capabilities. Thus, we decouple caption generation (by VLMs) from semantic alignment (by LLMs) to improve evaluation reliability. Unless otherwise specified, we adopt LLaVA-Video-7B~\cite{llavavideo} as our VLM model and Qwen2.5-7B-Instruct~\cite{yang2024qwen2} as LLM model.

\smallTitle{Video-Based Multi-Question Answering}. It is designed for evaluation dimensions where one salient concept is prominent and can be directly queried through video question answering (VQA). In this approach, we construct a series of complementary and sometimes redundant questions to reduce the risk of accidental errors and ask the VLM to perform direct VQA. The formalization is as follows: \begin{align} Answer&=\sum_i^N VQA(Q^i, V~|~S), \end{align} where \textit{Q} is a set of multiple questions. For example, in the \textit{Dynamic Attribute} dimension focusing on color changes, we may ask:
\textit{1. Initially, is the color of the river mostly blue? 2. Finally, is the color of the river mostly brown? 3. Does the color of the river change?}
The answer to each question is binary (“yes” or “no”), and scores are either averaged or awarded only if all responses are correct, depending on the dimension’s scoring scheme. This scheme is particularly effective for surface-level visual understanding, where modern VLMs can confidently answer targeted queries without requiring high-level semantic reasoning. Unless otherwise specified, we adopt LLaVA-Video-7B~\cite{llavavideo} as our VLM model.

 \begin{figure}[t]
  \centering
   \includegraphics[width=0.99\linewidth]{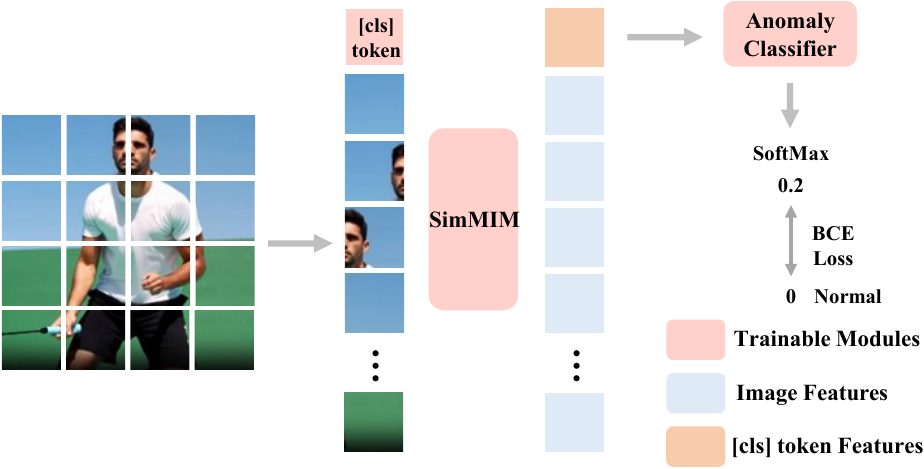}
   \caption{\textbf{Human Anatomy Detector Framework.} The [cls] token is used to aggregate information related to anomalies in the image. The human body, hand, and face detectors share the same pipeline with different data. The input is frame level and pre-detected.}
   \label{fig:anomaly_pipe}
\end{figure}

\noindent\textbf{1) Human Fidelity:}
We evaluate both the structural correctness and temporal consistency of human figures in generated videos. Structural issues commonly seen in current video generation models, such as sudden turns or the ``thousand-hand yoga'' effect, are considered. For temporal consistency, we assess the consistency of human identity and clothing across frames.


\begin{figure}[h]
  \centering
   \includegraphics[width=0.99\linewidth]{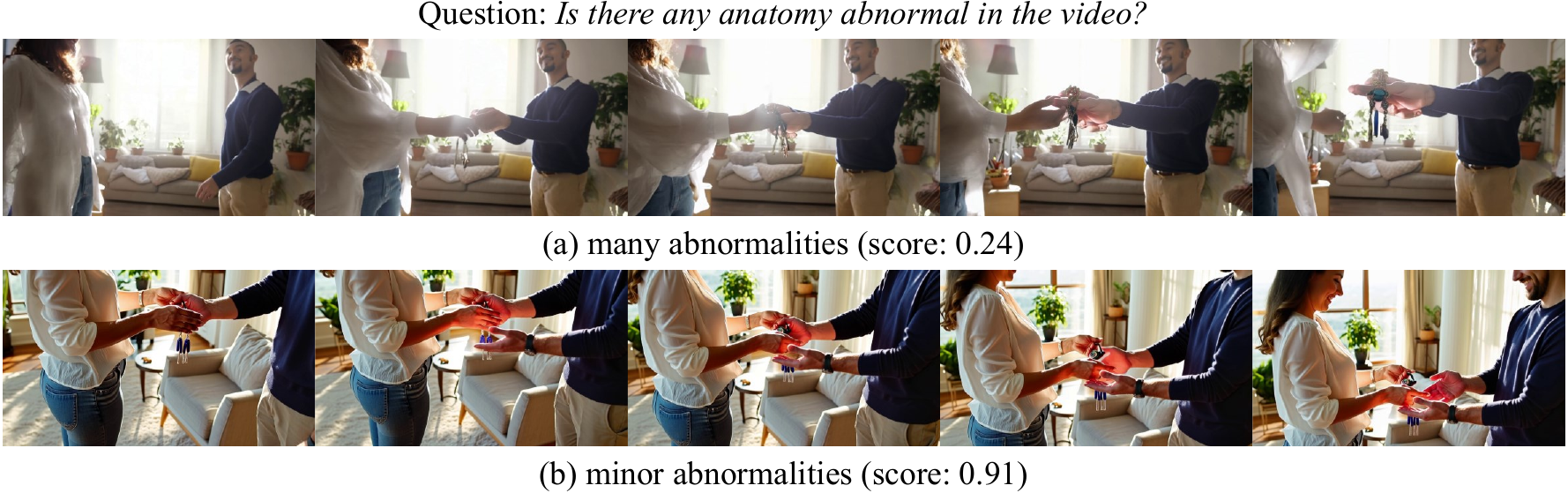}
   \caption{Visualization for \textit{Human Anatomy}.}
   \label{fig:ha}
   \vspace{-10pt}
\end{figure}

\textit{(1-a) Human Anatomy.} This dimension focuses on detecting potential anomalies in the appearance and structure of the human body in videos. We use a pre-trained ViT-base model \cite{xie2022simmim} to train three anomaly detection models, which will detect the human in each frame and judge the anomaly score of each human, targeting the human body, hands, and faces, respectively. The classification token (\texttt{cls\_token}) is passed through an MLP to produce binary outputs. We show the pipeline in Figure~\ref{fig:anomaly_pipe} and the dataset consists of two components:  
\begin{itemize}
    \item \textit{Real samples}: We collected approximately 1,000 real motion videos from the web and used the YOLO-World \cite{cheng2024yolo} detection model to extract pre-frame patches of human bodies, hands, and faces as positive samples.  
    \item \textit{Generated samples}: We generated around 1,000 videos related to human motion using CogVideo \cite{hong2022cogvideo, yang2024cogvideox} and HunyuanVideo \cite{kong2024hunyuanvideo}. These were processed with YOLO-World to extract image patches, followed by meticulous manual annotations for training. Additionally, negative samples were sampled from the HumanRefiner \cite{fang2024humanrefiner} dataset, focusing on human body, hands, and face. 
\end{itemize}
As a result, we collect 150k labeled real and generated human frame-level data. During inference, we first applied the YOLO-World model to detect all human instances in the input video. For each detected human region, we further detected hands and faces, extracting corresponding patches as inputs. A human instance is flagged as abnormal if any of the three models consider this human as an anomaly and the final score is the percentage of frames that are not flagged as abnormal.

\textit{(1-b) Human Temporal Consistency - Clothes.} 
Clothing consistency is assessed using video-based multi-question answering to assess whether outfits remain stable throughout video. Note that we do not use traditional methods such as feature similarity calculation to handle this problem based on two considerations: 1) Current models often generate characters with their clothing partially obscured by objects or with changes in visible body parts (e.g., the upper body transforming into the lower body). In such cases, traditional algorithms are unable to address these cross-temporal judgment issues. 2) LLaVA-video-7B has strong clothing color perception and a certain degree of cross-temporal memory capability. Therefore, we use it to evaluate this dimension.

\textit{(1-c) Human Temporal Consistency - Identity.} 
Identity consistency is evaluated by measuring facial feature similarity using ArcFace~\cite{deng2019arcface}, with face detection performed by RetinaFace~\cite{deng2020retinaface}. Specifically, we select the first frame of the video as the anchor by default, and all subsequent frames are compared to the first frame to calculate similarity. However, time periods where there are multiple people or no one present are not taken into consideration. \textbf{Note} that after a scene change, whether it is still the same person is taken into consideration, which represents a more challenging scenario.

\noindent\textbf{2) Creativity:}
We evaluate creativity by analyzing a model’s ability to generate diverse outputs and complex compositions beyond real-world constraints.

\begin{figure}[t]
  \centering
   \includegraphics[width=0.99\linewidth]{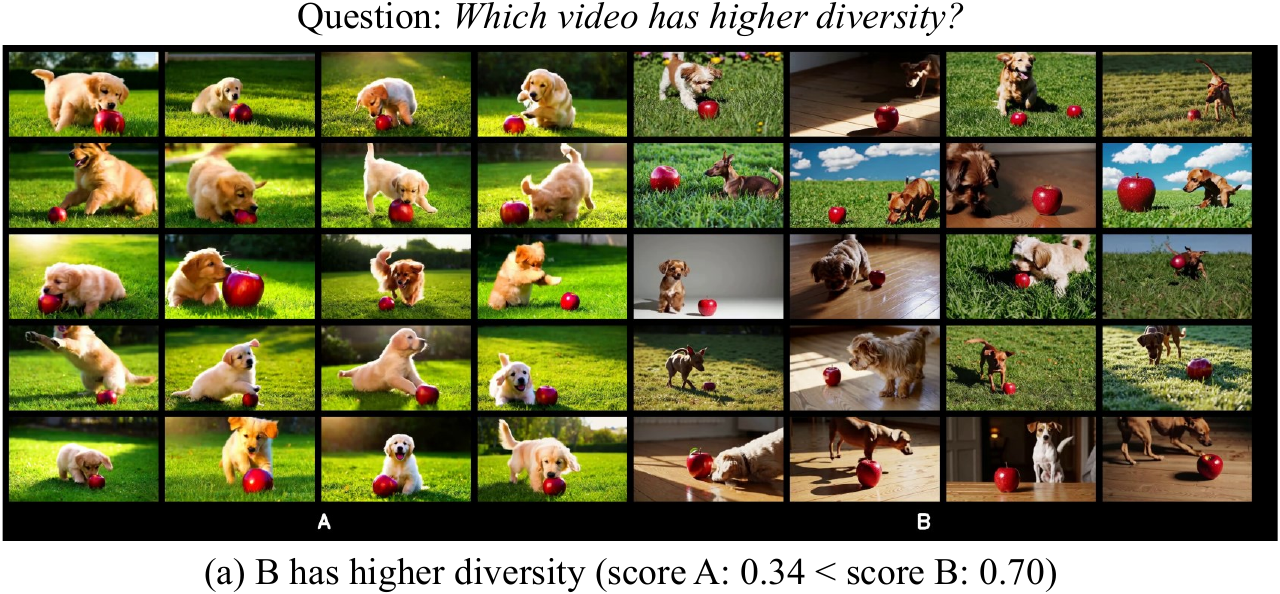}
   \caption{Visualization for \textit{Diversity}.}
   \label{fig:com}
   \vspace{-10pt}
\end{figure}

\textit{(2-a) Diversity.} 
Given a text prompt, we sample 20 videos from a model, and measure inter-sample variation using style and content diversity metrics, computed from pre-trained VGG-19~\cite{vgg} feature representations (\ie the metric is modified from~\cite{gatys2015neural}).

\textit{(2-b) Composition.} 
This dimension measures whether the model can generate novel and uncommon compositions and we assess it through three sub-dimensions: species combination, single-entity actions, and multi-entity interactions, using video-based multi-question answering pipeline. We additionally observe that pure VQA can not ensure robust results as current video generation models will generate separate species for the given prompt and VQA is a discrete process that could not consider the separate situations. So in this paper, we omit the cases that show more than one creature in the video by a pre-VQA question \textit{``Is there only one creature in the video?''}.

\noindent\textbf{3) Controllability:}
We evaluate a model’s ability to follow complex prompts and simulate dynamic changes during video generation. This dimension measures how accurately the model can render specific entities, events, content, and camera movements in response to detailed textual instructions.

\textit{(3-a) Entity - Dynamic Attribute.} 
We test whether models can modify attributes (\eg color, size, texture) mid-video using the \textit{video-based multi-question answering} pipeline. We take an example of questions here: \textit{``Initially, is the color of leaves mostly red?''; ``Finally, is the color of leaves mostly green?''; ``Does the color of leaves change?''}

\textit{(3-b) Entity - Dynamic Spatial Relationship.} 
We assess whether models accurately reposition objects in response to spatial instructions (\eg ``A dog is on the left of a sofa, then the dog runs to the front of the sofa.'') using \textit{video-based multi-question answering}. The question template is similar to \textit{Dynamic Attribute}.

\textit{(3-c) Event - Motion Order Understanding.} 
We evaluate whether models generate several actions or motions in the specified order. We use the \textit{text description alignment} pipeline to measure whether the generated motion sequence matches the text prompt.
Specifically, we first split the text prompt into two descriptions with action only in turn, then we obtain the video caption by VLM with an additional system prompt \textit{``Return the action order in video. Here is the template: `1. ; 2. .'''}. Then we assess whether both actions match the ground truth descriptions in turn; only when both actions match are they considered correct. This approach cleverly avoids the hallucination problem of LLM in order judgment.

\textit{(3-d) Event - Human Interaction.} 
We assess whether two humans can interact (\eg ``One person hands an object to another'') based on the text prompt using \textit{text description alignment}. The system prompt for VLM is \textit{``Describe the human interaction in the video, following the template as [a person xx to another person.]''}. After converting the description into a standardized format, the LLM will determine whether the standardized description matches the ground truth. 
However, the wrong number of people may lead to interaction hallucinations in VLMs, such as treating situations with only one person as interactions between two individuals, so we pre-filter the video with wrong human number with another \textit{text description alignment} pipeline. Specifically, we obtain the dense description of the video with system prompt \textit{``Describe the video in detail''} and ask the LLM to determine whether the number of people mentioned in the video description is exactly one.

\textit{(3-e) Content - Complex Plot.} 
We test a model’s ability to construct multi-scene narratives from prompts describing multi-stage events (\eg a five-act story with 150+ words) to meet the demands of future movie-level video generation.  We utilize \textit{text description alignment} to evaluate it. Specifically, we first manually split and summarize the long text prompt into 5 plot descriptions as ground truth. Then we feed the video into VLM to obtain 5 video captions with the system prompt \textit{``Return the plot in video. Here is the template: [1. ; 2. ; 3. ; 4. ; 5. .]''} (\ie note that when dealing with long videos at the minute level, VLMs may make errors in their point-by-point summarization. Therefore, we additionally employ an LLM for post-processing verification to correct captions with incorrect numbering). Finally, we sequentially match the corresponding text and captions in the order of the plot. If the LLM determines that a plot element appears in the caption, we proceed to evaluate the next plot element; otherwise, the evaluation stops. We discretize the plot elements to calculate the accurate score, simulating the effect of LLM scoring.

\begin{figure}[t]
  \centering
   \includegraphics[width=0.99\linewidth]{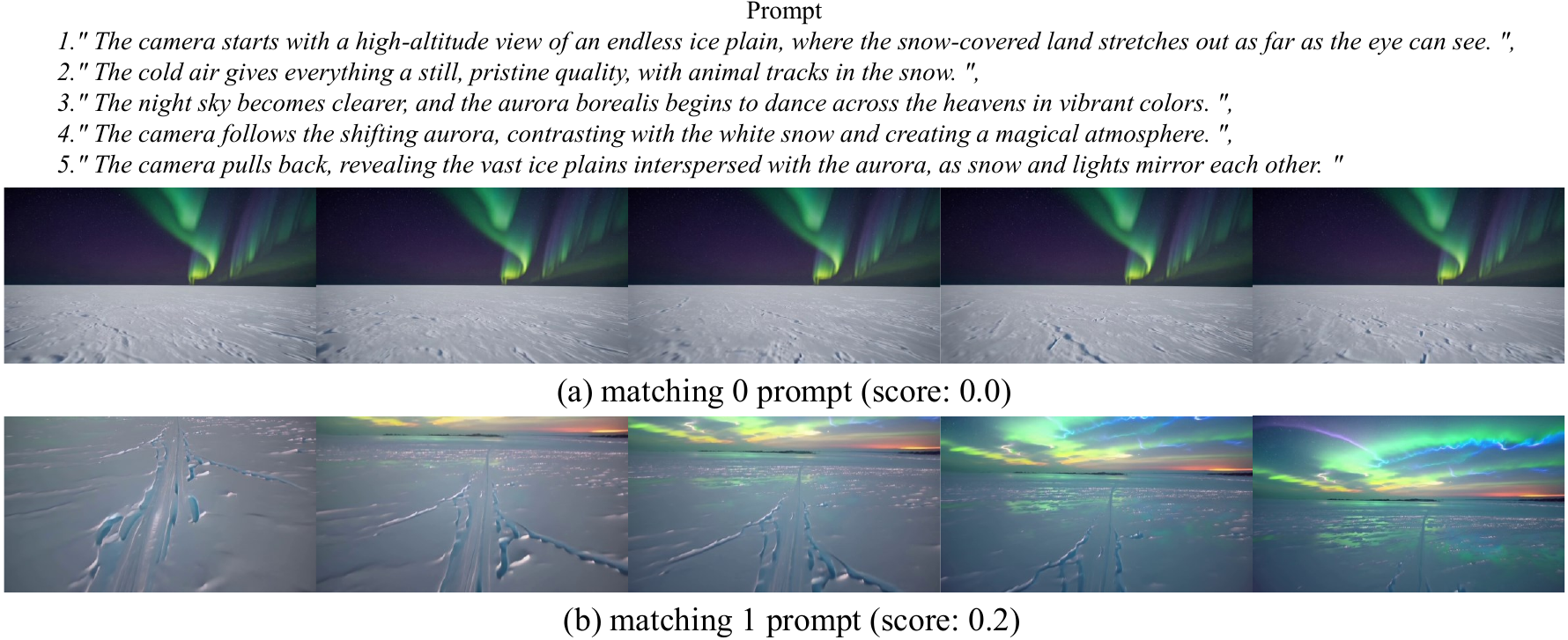}
   \caption{Visualization for \textit{Complex Landscape}.}
   \label{fig:cl}
   \vspace{-10pt}
\end{figure}

\textit{(3-f) Content - Complex Landscape.} 
We evaluate whether models faithfully follow long-form landscape descriptions (150+ words) that include multiple scene transitions driven by camera movements. We assess adherence using \textit{text description alignment} with landscape-specific system prompt, similar to \textit{Complex Plot}.

\textit{(3-g) Camera Motion.} 
We evaluate whether models can generate specified camera movements. We extend VBench++~\cite{huang2024vbench++}'s camera motion taxonomy to nine types, adding ``Orbit'' and ``Oblique shot, airborne dolly movement.'' The generated camera motion is assessed via point tracking with CoTracker-v2~\cite{karaev2024cotracker} and carefully tailored heuristics (\eg different camera motions exhibit distinct behaviors at different points. For example, \textit{pan left} results in all points shifting to the right.).

\noindent\textbf{4) Physics:}
The \textit{Physics} dimension evaluates whether video generation models adhere to real-world physical principles. We assess two key areas: \textit{1) State Change}, which examines how well models simulate mechanical, thermal, and material transformations, and \textit{2) Geometry}, which evaluates the 3D consistency of objects and scenes across different frames. We extend prior benchmarks (\eg PhyGenBench~\cite{phygenbench}) by increasing physics scenario difficulty and significantly improving evaluation accuracy with tailored pipelines.

\begin{figure}[t]
  \centering
   \includegraphics[width=0.99\linewidth]{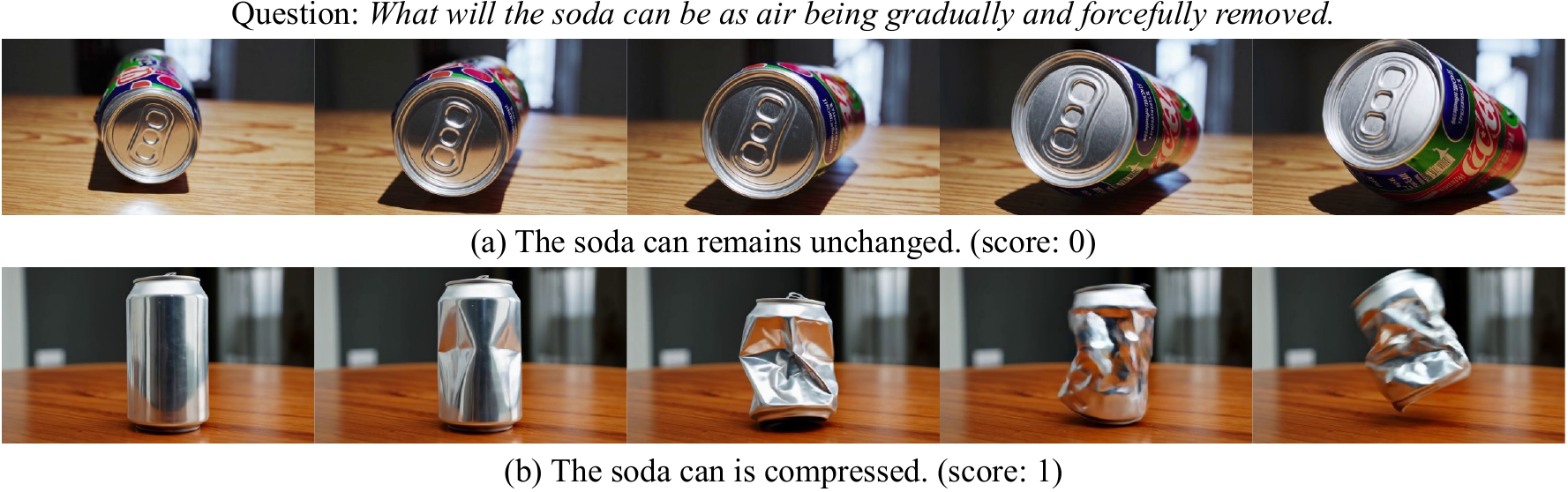}
   \caption{Visualization for \textit{Mechanics}.}
   \label{fig:phy}
   \vspace{-10pt}
\end{figure}

\textit{(4-a) State Change - Mechanics.} 
We evaluate whether models follow basic mechanical principles such as gravity, buoyancy, and stress. This is done using \textit{video-based multi-question answering}. Unlike PhyGenBench, which relies on abstract physics concepts, we prompt GPT-4o to generate explicit \textit{visual} descriptions of expected physical behavior (\eg unlike PhyGenBench that uses terminologies to describe the physical phenomena \textit{``The liquid's behavior aligns with the microgravity environment, floating freely, and forming natural blobs without noticeable distortion.''}, we describe the physical phenomena's visual results \textit{``The liquid floating, and forming blobs''})). Notably, the GPT-4o-generated descriptions are based solely on the text prompt, \textit{remain fixed during evaluation, and do not compromise evaluation reproducibility}.
To ensure a focused assessment of state changes, we also apply a pre-filtering step to exclude cases where the initial state of the generated video does not align with the prompt.

\textit{(4-b) State Change - Thermotics.} 
We evaluate how well the models simulate state transitions such as vaporization, liquefaction, and sublimation. To increase complexity, we introduce temperature-specific prompts (\eg ``A timelapse captures dry ice transforming at -90°C''). Evaluation follows the same \textit{video-based multi-question answering} approach used in the \textit{Mechanics} dimension.

\textit{(4-c) State Change - Material.} 
We evaluate whether the models correctly depict color mixing, hardness, combustion, and solubility. Similar to \textit{Mechanics} and \textit{Thermotics}, we use \textit{video-based multi-question answering}, testing whether generated videos properly follow material properties.

\textit{(4-d) Geometry - Multi-View Consistency.} 
Geometry is a critical aspect for 3D/4D video generation, ensuring that generated entities and scenes maintain structural consistency when viewed from different angles or as the camera moves. However, we do not have access to explicit 3D ground truth of the generated videos, making direct 3D validation infeasible. Instead, we assess multi-view consistency using two complementary metrics: \textit{1) Feature Matching Stability} – Measures how well objects retain their geometric consistency across frames, inspired by \cite{li2024sora}, and \textit{2) Camera Motion Speed} – Accounts for the effect of motion strengths on feature stability, ensuring fair cross-model comparisons. Specifically, we follow  \cite{li2024sora} to extract frame-level keypoint features using SIFT~\cite{lowe2004sift}, efficiently match points across frames with FLANN~\cite{muja2009fast}, and eliminate incorrect matches using RANSAC~\cite{fischler1981random}. We then estimate camera motion speed using RAFT~\cite{teed2020raft}, and adjust feature-matching frame intervals based on the motion strength and the video frame rate (\ie FPS and camera motion speed will influence the score of feature matching).

\noindent\textbf{5) Commonsense:}
We assess commonsense reasoning in video generation models across two key aspects: \textit{1) Motion Rationality}, evaluating whether generated motions are physically plausible and correctly executed, and \textit{2) Instance Preservation}, ensuring that abnormal entity states (entity sudden merging, splitting, appearing, and disappearing) do not exist.

\begin{figure}[t]
  \centering
   \includegraphics[width=0.99\linewidth]{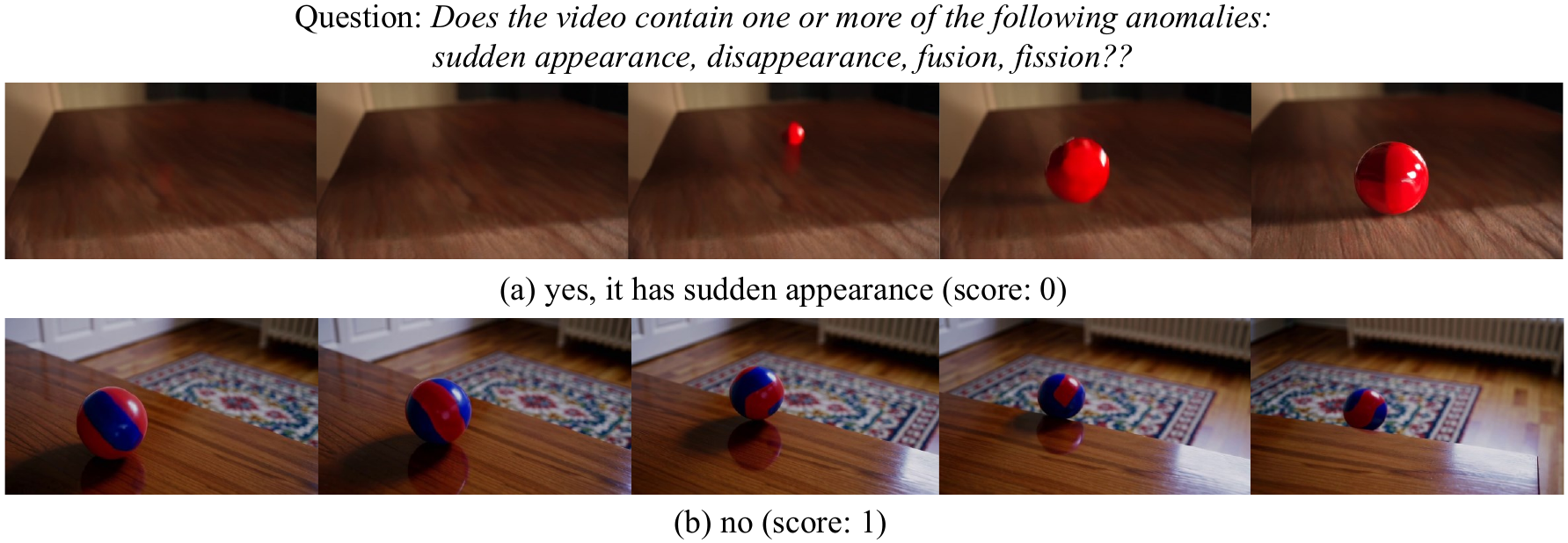}
   \caption{Visualization for \textit{Instance Preservation}.}
   \label{fig:ip}
   \vspace{-10pt}
\end{figure}

\begin{figure*}[t!]
    \centering
    \begin{minipage}{0.60\textwidth}
        \centering
        \vspace{-5pt}
        \includegraphics[width=0.99\linewidth, height=0.4\textwidth]{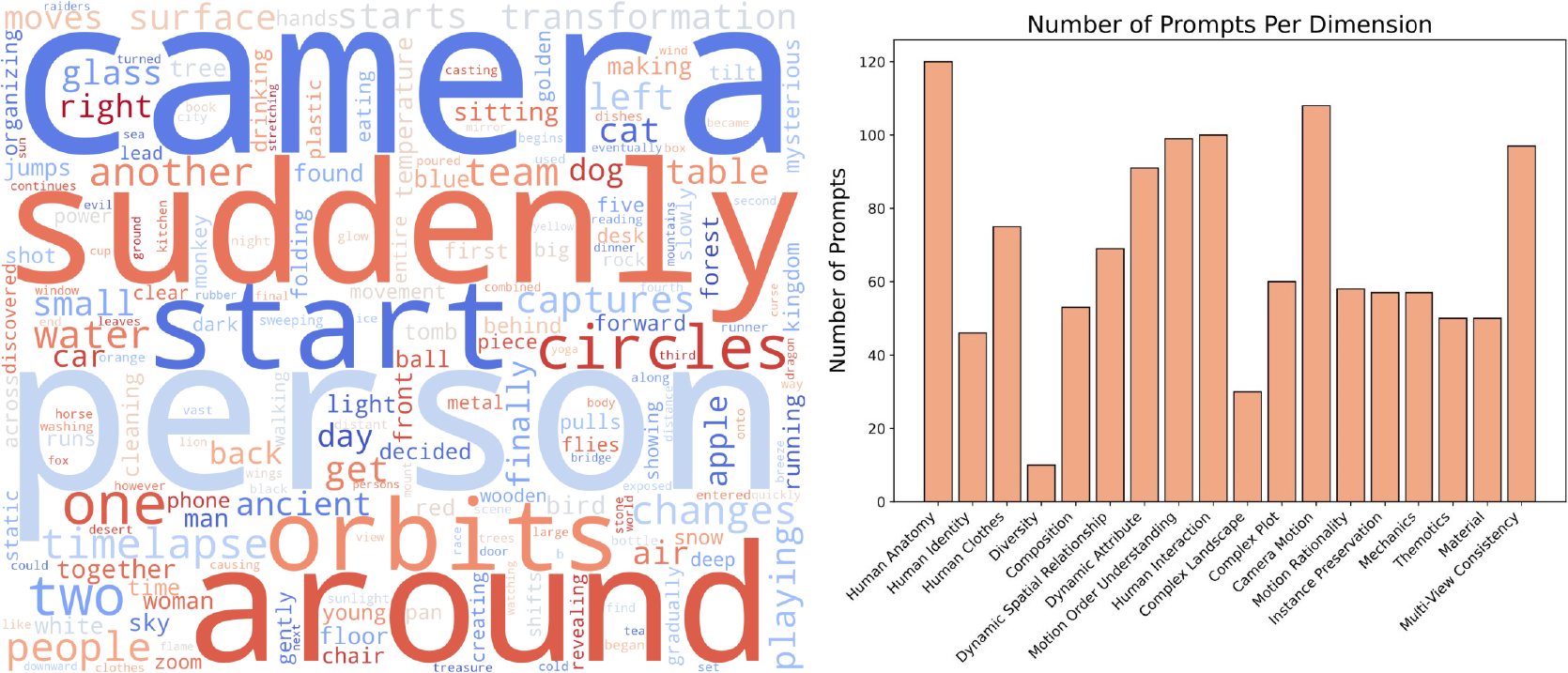}
        \caption{\textbf{Overview of Prompt Suite Statistics.} \textbf{\textit{Left:}} distribution of words in the prompt suites.  \textbf{\textit{Right:}} number of prompts per evaluation dimension.}
        \label{fig:prompt_suite_statistics}
    \end{minipage}\hspace{0.01\textwidth}
    \begin{minipage}{0.38\textwidth}
        \centering
        \vspace{-5pt}
        \includegraphics[width=0.99\linewidth]{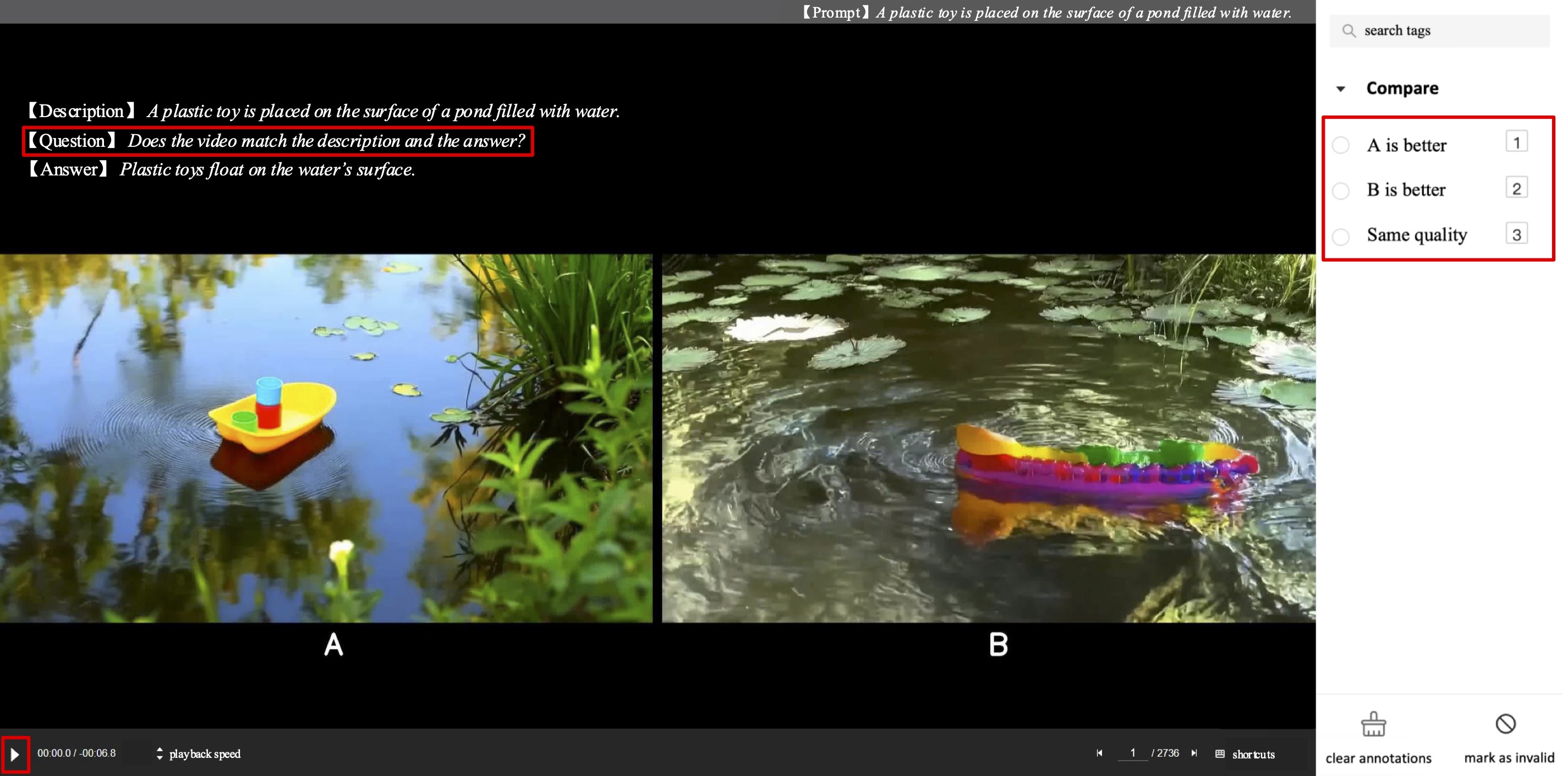} 
        \caption{\textbf{Interface for Human Preference Annotation}. \textbf{\textit{Top:}} Question descriptions. \textbf{\textit{Right:}} Choices available to annotators. \textbf{\textit{Bottom left:}} Controls for stopping and playback.}
        \label{fig:fig_paper_interface}
    \end{minipage}
\vspace{-10pt}
\end{figure*}

\textit{(5-a) Motion Rationality.} 
We evaluate whether the generated motion leads to the correct real-world consequences. A prevalent issue in video generation is the presence of fake motions that visually appear correct but lack expected effects on the environment.
For instance: 1) Fake eating: A person bites into food, but the food remains unchanged. 2) Fake walking: A character moves their legs, but does not actually progress forward. and 3) Fake cutting: A knife moves through an object, but the object does not split.
To address this, we adopt the \textit{video-based multi-question answering} pipeline, specifically to ensure the motion's impact is checked explicitly (\eg ``After slurping, is the amount of noodles visibly reduced?'').
Redundant questioning (\eg 1. ``Does the person appear to be slurping noodles?'', 2. ``Is the person's mouth in contact with the noodles?'', 3. ``Is the bowl of noodles remaining after slurping?'') helps filter out false positives and accidental misinterpretations.

\textit{(5-b) Instance Preservation.} 
Video generative models frequently generate videos with unnatural entity merging, duplication, or disappearance, particularly during large-motion sequences or object interactions. As current VLMs could not capture this clip-level or even frame-level abnormal (\ie Unnatural merging can appear between adjacent frames, despite both frames being normal on their own). We tune a clip-level entity abnormal detector based on Qwen2.5-VL-3B-Instruct~\cite{bai2025qwen25vl}. Specifically, we use a multiple-choice question format to ask the VLM to determine whether a given clip contains any of the aforementioned anomalies (A for ``yes'' and B for ``no''). Only the all of the clips in a video is normal will be considered as normal (score 1), otherwise 0. 

In terms of training data, we first collected a set of videos that align with our prompt theme as normal samples. We also sampled several generative models (CogVideo, HunyuanVideo, Wan2.1, StepVideo), and manually annotated anomalous samples. To reduce the gap between virtual and real data, both virtual and real normal data is used for training and  we further apply LoRA fine-tuning to maintain the model's generalization ability as much as possible. 

\subsection{Prompt Suite}
\label{subsec:prompt_suite}

The VBench-2.0 Prompt Suite is designed to be compact yet representative. Given the increasing computational cost of video sampling, especially for longer and higher-resolution videos (\eg HunyuanVideo and CogVideoX-1.5 taking over five minutes per sample on 8×A100 GPUs), we strategically limit the number of test cases to reduct sampling costs during evaluation, while ensuring coverage across diverse evaluation dimensions and content scenarios. Figure~\ref{fig:prompt_suite_statistics} visualizes the prompt distributions.

\smallTitle{Tailored Prompts for Each Dimension.} 
For each evaluation dimension in VBench-2.0, we carefully construct a suite of approximately 70 prompts, specifically tailored to probe the model’s capabilities in that dimension. Prompts are designed to systematically analyze the core ability being tested.
For example, in the \textit{Multi-View Consistency} dimension, we evaluate both object-level and scene-level prompts, ensuring that models are tested across varying spatial structures. In the \textit{Composition} dimension, we systematically divide prompts into species combination, single-entity action, and multi-entity tasks, covering different levels of creativity and compositional reasoning. The \textit{Physics} dimension follows a structured approach, incorporating \textit{Mechanics} (\eg gravity, buoyancy, stress), \textit{Thermotics} (\eg vaporization, freezing), and \textit{Material} properties (\eg color mixing, solubility) to comprehensively assess adherence to physical laws. To further challenge model reasoning, additional constraints, such as temperature-specific prompts in \textit{Thermotics}, require models to demonstrate a deeper understanding in how temperature affect thermotics beyond simple pattern matching.

\smallTitle{Ensuring Disentangled Evaluation.}
Prompts are designed to eliminate confounding factors and ensure focused assessment in the dimension of interest. For example, in \textit{Dynamic Spatial Relationship} and \textit{Dynamic Attribute}, only one entity is allowed to move or changed, ensuring the test solely assesses positioning and attribute rather than taking the irrelevant ability of multi-object interactions into account.

\smallTitle{Evaluation Robustness.} 
To improve the robustness of evaluation, we design the actions and events in each dimension's prompts to be explicitly recognizable by VLMs and LLMs. In the \textit{Human Interaction} dimension, we ensure that prompts include physical contact interactions (\eg ``A person shakes hands with another'') rather than ambiguous social scenarios that are harder to verify visually (\eg ``Two people are having a picnic''). For \textit{Motion Rationality}, prompts are designed to ensure that the counterexamples involve visually observable outcomes, such as fake eating (food remains unchanged), fake walking (not moving forward), or fake cutting (objects remain unaltered), rather than examples like lip-syncing, which are difficult to conclusively assess through visual observation alone.

By following these structured design principles, VBench-2.0 provides a \textit{compact, diverse, and reliable} benchmark for evaluating video generation models across a diverse range of real-world and abstract scenarios. 

\subsection{Human Preference Annotation}  
\label{subsec:human_preference_annotation}
Following the approach of VBench~\cite{huang2024vbench, huang2024vbench++}, we conduct large-scale human preference labeling on generated videos to validate the alignment between VBench-2.0's evaluation and human perception across all evaluation dimensions. The collected human annotations also serve as a valuable resource for future research on fine-tuning generation and evaluation models to better reflect human judgments.

\noindent\textbf{Data Preparation.} There are two types of annotation formats in VBench-2.0.

The first type follows VBench and the human annotators are tasked to select a preferred video from two generated videos based on specific criteria.
Given a text prompt $p_{i}$ and four video generation models $\{A, B, C, D\}$, we generate a set of videos, forming a ``group'' $G_{i,j}=\{V_{i,A,j}, V_{i,B,j}, V_{i,C,j}, V_{i,D,j}\}$. For each prompt $p_{i}$, we sample five such groups $\{G_{i,0}, G_{i,1}, G_{i,2}, G_{i,3}, G_{i,4}\}$ and construct pairwise comparisons: 
$(V_{A}, V_{B})$, $(V_{A}, V_{C})$, $(V_{A}, V_{D})$, $(V_{B}, V_{C})$, $(V_{B}, V_{D})$, $(V_{C}, V_{D})$.
Human annotators are asked to select their preferred video for each pair. To ensure unbiased annotations, the video order within each pair is randomized. 

The second type might involve two groups of videos, and the dimension in evaluation is related to the content distribution in these two groups of videos. In this type, the pairwise construction is the same as the first type while each $V_i$ contains 20 videos generated by single prompt $p_i$.

\noindent\textbf{Labeling Instructions.} 
The annotation process follows VBench but incorporates refinements in interface design and evaluation methodology. Since VBench-2.0 introduces multiple dimensions with detailed multi-question evaluations and long text prompts (some exceeding 150 words), we enhance the interface for improved readability and efficiency. Instead of displaying extensive text in video titles, we sequentially list all key annotation instructions directly within the interface. This structured layout allows annotators to efficiently reference the necessary details while conducting evaluations.

\noindent\textbf{Win Ratio.} Given human annotations, we calculate the win ratio for each model. During pairwise comparisons, if a model's video is preferred, it scores 1, while the other model scores 0. In case of a tie, both models score 0.5. The win ratio for each model is computed as the total score divided by the number of pairwise comparisons. We show the result in Table~\ref{tab:win_ratio}.

\noindent\textbf{Quality Assurance.} 
The annotation process maintains the rigorous quality control measures established in VBench while further refining the review criteria. We randomly sample 20\% of the annotated pairs for verification, with a required success rate of 95\%.

To quantify the annotation effort across all 18 evaluation dimensions, we account for the cumulative time spent on documentation drafting, initial trials, formal annotation, and re-annotation. In total, the process required 15.75 hours of individual effort and 284 hours across 18 annotators, reflecting the scale of the task and our commitment to ensuring high-quality human preference annotations. Most evaluation dimensions go through 2 rounds of trial labeling, official labeling, and post-labeling verification.

\begin{table*}[t!]
\centering
\begin{center}
\caption{\textbf{VBench-2.0 Evaluation Results per Dimension.} This table presents evaluation results for four recent state-of-the-art video generation models across all 18 VBench-2.0 dimensions. A higher score indicates better performance in the corresponding dimension.
}
\vspace{-5pt}
\resizebox{0.98\linewidth}{!}{
\begin{tabular}{c|c|c|c|c|c|c|c|c|c}
\hline
\textbf{Models}   & \textbf{\Centerstack{Human\\Anatomy}} & \textbf{\Centerstack{Human\\Clothes}} & 
\textbf{\Centerstack{Human\\Identity}} & \textbf{\Centerstack{Composition}} & \textbf{\Centerstack{Diversity}} & \textbf{Mechanics} & \textbf{Material} & \textbf{Thermotics} & \textbf{\Centerstack{Multi-view\\Consistency}}\\ \hline
HunyuanVideo~\cite{kong2024hunyuanvideo}        & \textbf{88.58\%} & 82.97\% & 75.67\% & 43.96\% & 39.73\% & 76.09\% & 64.37\% & 56.52\% & 43.80\% \\
CogVideoX-1.5~\cite{yang2024cogvideox}  & 59.72\% & 87.18\% & 69.51\% & 44.70\% & 42.61\% & \textbf{80.80\%} & \textbf{83.19\%} & \textbf{67.13\%} & 21.79\% \\ 
Sora~\cite{sora} & 86.45\% & \textbf{98.15\%} & \textbf{78.57\%} & \textbf{53.65\%} & \textbf{67.48\%} & 62.22\% & 64.94\% & 43.36\% & 58.22\% \\ 
Kling 1.6~\cite{kling}  & 86.99\% & 91.75\% & 71.95\% & 43.89\% & 53.26\% & 65.55\% & 68.00\% & 59.46\% & \textbf{64.38\%} \\ \hline

\textbf{Models} & \textbf{\Centerstack{Dynamic Spatial\\Relationship}} &  \textbf{\Centerstack{Dynamic\\Attribute}} & \textbf{\Centerstack{Motion Order\\Understanding}} & \textbf{\Centerstack{Human\\Interaction}} & \textbf{\Centerstack{Complex\\Landscape}} & \textbf{\Centerstack{Complex\\Plot}} & \textbf{\Centerstack{Camera\\Motion}} & \textbf{\Centerstack{Motion\\Rationality}} & \textbf{\Centerstack{Instance\\Preservation}} \\
\hline
HunyuanVideo~\cite{kong2024hunyuanvideo} & \textbf{21.26\%} & 22.71\% & 26.94\% & 66.67\% & 18.89\% & 9.78\% & 33.95\% & 34.48\% & 92.40\% \\ 
CogVideoX-1.5~\cite{yang2024cogvideox}  &19.32\% & \textbf{24.18\%} & 26.60\% & \textbf{72.33\%} & \textbf{20.00\%} & \textbf{11.21\%} & 33.33\% & 33.91\% & 82.46\% \\ 
Sora~\cite{sora} & 19.81\% & 8.06\% & 15.15\% & 58.00\% & 15.33\% & 11.11\% & 27.16\% & 34.48\% & \textbf{94.15\%} \\
Kling 1.6~\cite{kling}  & 20.77\% & 19.41\% & \textbf{29.29\%} & 72.00\% & 17.33\% & 10.83\% & \textbf{61.73\%} & \textbf{38.51\%} & 92.40\%\\ \hline
\end{tabular}
}
\label{tab:raw_metrics}
\vspace{-10pt}
\end{center}
\end{table*}

\section{Experiments}

\subsection{Video Generation Models in Evaluation}
\label{sec:setup}
To assess our benchmark against recent advancements, we utilize four models for comparison, with more to be included as they become open-sourced. We show the basical information of the four models in Table~\ref{tab:model_info} and  introduce the detailed implementation and a unified prompt refiner to pre-process the text prompt as follows.

\noindent\textbf{Prompt Refiner.} As video generation models continue to improve their understanding of text and support longer inputs, many methods have begun adopting prompt rewriting techniques to refine text input. To ensure fair comparisons and higher quality video in this paper, we employ a modified version of the original Prompt Refiner from CogVideoX to uniformly process all input prompts. We tested the following four models and found that except for Sora, our Prompt Refiner consistently result in higher-quality videos with better alignment to the input text. We hypothesize that Sora may have an embedded Prompt Refiner, which could explain its performance. Therefore, we applied our Prompt Refiner to \textbf{all dimensions} of the other three models. 

\noindent\textbf{Kling 1.6~\cite{kling}.} We adopt the standard API version of Kling 1.6. For each prompt, we sample 241 continuous frames of size 720$\times$1280 at 24 frames per second (FPS). For the classifier-free guidance (cfg) scale, we used the official default value of 0.5.

\noindent\textbf{Sora-480p~\cite{sora}.} Sora represents one of the earliest large-scale video generation models to emerge. For each prompt, we directly perform sampling on the official website, containing 150 continuous frames of size 480$\times$854 at 30 FPS (\ie, the low-resolution version of Sora).

\noindent\textbf{HunyuanVideo~\cite{kong2024hunyuanvideo}.} HunyuanVideo is a powerful open-sourced video generation model. It is equipped with a pre-trained Multimodal Large Language Model (MLLM)~\cite{sun2024hunyuan} to better understand text prompts, supporting a large maximum token length and incorporating a dual-branch diffusion transformer generative architecture, which produces visually convincing results. We use the best version of HunyuanVideo with no classifier-free guidance. For each prompt, we sample 129 continuous frames of size 720$\times$1280 at 24 FPS. All of the hyper-parameters are followed the default value in the official inference code. The initial random seed is set to 42 during sampling.

\noindent\textbf{CogVideoX-1.5~\cite{yang2024cogvideox}.} CogVideoX-1.5 is a transformer-based video generation model, which is equipped with a 3D causal VAE to compress the spatial and temporal dimensions and an expert transformer generative architecture to achieve better text-video alignment. We use the best version of it (5B) to evaluate. For each prompt, we sample 161 continuous frames of size 768$\times$1360 at 16 FPS. The initial random seed is also set to 42 for a fair comparison.

\begin{table}[t]
\centering
\caption{\textbf{Information on Evaluated Models.}}
\vspace{-3pt}
\resizebox{1\linewidth}{!}{
\begin{tabular}{c|ccc}
\hline
 \textbf{\Centerstack{Model Name}} &
\textbf{\Centerstack{Video Length}} 
& \textbf{\Centerstack{Per-Frame Resolution}} & 
\textbf{\Centerstack{Frame Rate (FPS)}} 
\\ \hline
HunyuanVideo~\cite{kong2024hunyuanvideo}    & 5.3s & 720$\times$1280 & 24  \\
CogVideoX-1.5~\cite{yang2024cogvideox}    & 10.1s & 768$\times$1360 & 16  \\ 
Sora-480p~\cite{sora}    & 5.0s & 480$\times$854 & 30  \\
Kling 1.6~\cite{kling}    & 10.0s & 720$\times$1280 & 24  \\ \hline
\end{tabular}
}
\label{tab:model_info}
\vspace{-10pt}
\end{table}

\begin{table*}[!htb]
\vspace{5pt}
\centering
\caption{
\textbf{Human Alignment of VBench-2.0 Evaluation Methods.} For each evaluation dimension and each video generative model, we report ``\textit{VBench-2.0 Win Ratios (left) / Human Win Ratios (right)}''. The results demonstrate that our evaluation metrics closely align with human perception across all dimensions.}
\resizebox{1\linewidth}{!}{
\begin{tabular}{c|c|c|c|c|c|c|c|c|c}
\hline
\textbf{Models}   & \textbf{\Centerstack{Human\\Anatomy}} & \textbf{\Centerstack{Human\\Clothes}} & 
\textbf{\Centerstack{Human\\Identity}} & \textbf{\Centerstack{Composition}} & \textbf{\Centerstack{Diversity}} & \textbf{Mechanics} & \textbf{Material} & \textbf{Thermotics} & \textbf{\Centerstack{Multi-View\\Consistency}}\\ \hline%
HunyuanVideo~\cite{kong2024hunyuanvideo}  & \textbf{67.73\%} / 58.73\% & 44.49\% / 45.67\% & 52.60\% / 58.19\% & 46.02\% / 40.25\% & 16.67\% / 36.67\% & 53.30\% / 50.64\% & 45.70\% / 45.49\% & 50.13\% / 46.08\% & 52.83\% / 50.39\% \\ 
CogVideoX-1.5~\cite{yang2024cogvideox}  & 13.10\% / 5.28\% & 48.12\% / 44.28\% & 36.81\% / 20.62\% & 50.00\% / 46.23\% & 26.67\% / 28.33\% & \textbf{57.57\%} / \textbf{51.20\%} & \textbf{59.43\%} / \textbf{57.49\%} & \textbf{56.69\%} / \textbf{58.51\%} & 8.77\% / 18.07\% \\ 
Sora~\cite{sora} & 60.05\% / \textbf{69.71\%} & \textbf{55.61\%} / \textbf{57.15\%} & 53.88\% / 56.25\% & \textbf{56.08\%} / \textbf{59.96\%} & \textbf{93.33\%} / \textbf{83.33\%} & 43.47\% / 49.39\% & 47.38\% / 50.79\% & 40.37\% / 41.19\% & 64.10\% / 63.94\% \\ 
Kling 1.6~\cite{kling}  & 59.12\% / 66.36\% & 51.18\% / 52.86\% & \textbf{56.48\%} / \textbf{63.05\%} & 47.90\% / 53.56\% & 63.33\% / 51.67\% & 45.56\% / 48.78\% & 46.65\% / 45.90\% & 52.74\% / 54.02\% & \textbf{80.00\%} / \textbf{65.98\%} \\ \hline
Correlation  & 95.46\% & 90.89\% & 99.46\% & 81.70\% & 94.53\% & 93.56\% & 93.71\% & 93.86\% & 98.44\% \\ \hline
\hline
\textbf{Models}  & \textbf{\Centerstack{Dynamic Spatial\\Relationship}} &  \textbf{\Centerstack{Dynamic\\Attribute}} & \textbf{\Centerstack{Motion Order\\Understanding}} & \textbf{\Centerstack{Human\\Interaction}} & \textbf{\Centerstack{Complex\\Landscape}} & \textbf{\Centerstack{Complex\\Plot}} & \textbf{\Centerstack{Camera\\Motion}} & \textbf{\Centerstack{Motion\\Rationality}} & \textbf{\Centerstack{Instance\\Preservation}} \\
\hline
HunyuanVideo~\cite{kong2024hunyuanvideo} & \textbf{50.64\%} / \textbf{51.37\%} & 52.75\% / \textbf{55.07\%} & 51.63\% / 56.35\% & 49.61\% / 52.45\% & 52.41\% / \textbf{60.04\%} & 46.85\% / 30.46\% & 46.60\% / 49.28\% & 49.43\% / 52.68\% & 51.36\% / 50.88\% \\ 
CogVideoX-1.5~\cite{yang2024cogvideox}  & 49.36\% / 48.07\% & \textbf{53.72\%} / 54.70\% & 51.40\% / 52.13\% & \textbf{53.39\% }/ 51.34\% & \textbf{53.89\%} / 54.46\% & \textbf{52.31\%} / \textbf{65.37\%} & 46.19\% / 44.08\% & 49.04\% / 42.24\% & 44.74\% / 43.27\% \\ 
Sora~\cite{sora} & 49.68\% / 49.60\% & 42.98\% / 43.16\% & 43.77\% / 30.81\% & 43.83\% / 43.49\% & 42.04\% / 35.87\% & 51.30\% / 49.35\% & 42.08\% / 42.64\% & 49.43\% / 44.83\% & \textbf{ 52.53\%} / \textbf{53.80}\% \\ 
Kling 1.6~\cite{kling}  & 50.32\% / 50.97\% & 50.55\% / 47.07\% & \textbf{ 53.20\%} / \textbf{60.73\%} & 53.17\% / \textbf{52.73\%} & 51.67\% / 49.63\% & 49.54\% / 54.81\% & \textbf{65.12\%} / \textbf{63.99\%} & \textbf{52.11\%} / \textbf{60.25\%} & 51.36\% / 52.05\%\\ \hline
Correlation  & 97.36\% & 90.95\% & 99.31\% & 89.28\% & 91.82\% & 89.59\% & 97.95\% & 87.97\% & 99.11\% \\
\hline
\end{tabular}
}
\label{tab:win_ratio}
\end{table*}

\subsection{VBench-2.0 Evaluation Results}
 For each sub-ability dimension, videos were generated using the models based on the corresponding prompt suite described in Section~\ref{subsec:prompt_suite}. The evaluation method introduced in Section~\ref{subsec:evaluation_dimension_suite} is then applied to obtain numerical scores between 0 and 1, where a higher value indicate relatively stronger performance in that dimension. The evaluation results of the four video generative models are summarized in Table~\ref{tab:raw_metrics} and visualized in Figure~\ref{fig:fig_paper_radar_big}. Additionally, for each of the five major dimensions, we visualize the evaluation results of one model pair to demonstrate the accuracy of our method in Figure~\ref{fig:cl} (\textit{Complex Landscape} in \textit{Controllability}), Figure~\ref{fig:com} (\textit{Diversity} in \textit{Creativity}), Figure~\ref{fig:ha} (\textit{Human Anatomy} in \textit{Human Fidelity}), Figure~\ref{fig:phy} (\textit{Mechanics} in \textit{Physics}) and Figure~\ref{fig:ip} (\textit{Instance Preservation} in \textit{Commonsense}).

 \begin{figure*}[t]
    \vspace{-10pt}
  \centering
   \includegraphics[width=0.99\linewidth]{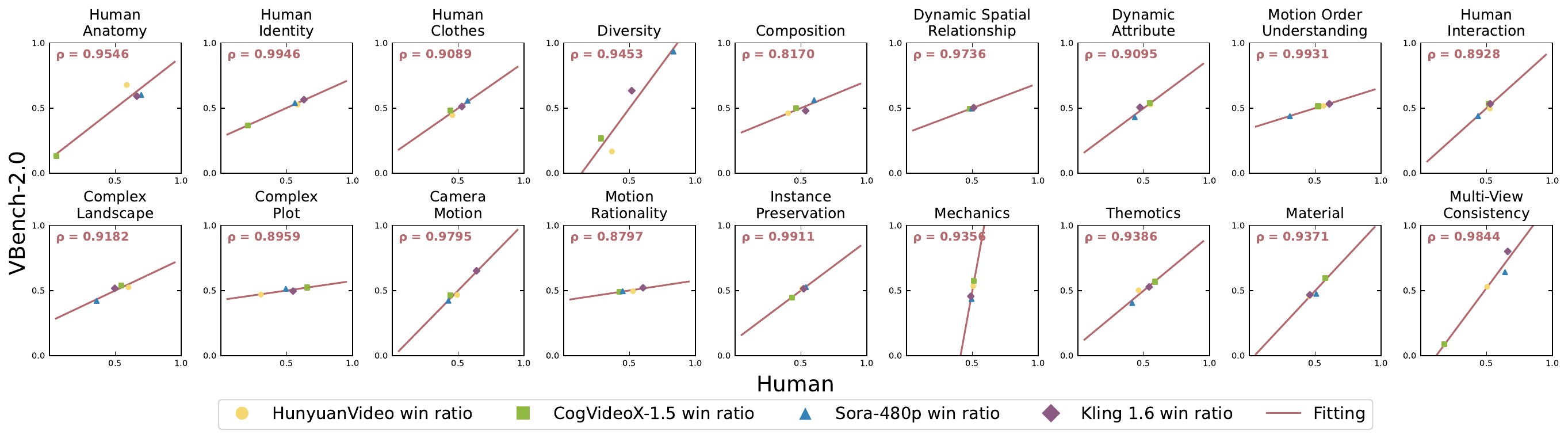}
   \caption{\textbf{Human Alignment of VBench-2.0 Evaluation.} Each plot represents the alignment verification for a specific VBench-2.0 dimension. In each plot, a dot corresponds to the human preference win ratio (horizontal axis) and the VBench-2.0 evaluation win ratio (vertical axis) for a given video generation model. A linear fit is applied to visualize the correlation, and Spearman’s correlation coefficient ($\rho$) is computed for each dimension. Experiments show that VBench-2.0 evaluations closely align with human judgement in all dimensions. 
   }
   \label{fig:win_ratio_dot}
   \vspace{-10pt}
\end{figure*}

\subsection{Human Alignment of VBench-2.0}
To ensure that VBench-2.0's evaluation aligns closely with human judgment across all evaluation dimensions, we conducted human preference labeling on a large set of generated videos, following the approach of VBench~\cite{huang2024vbench, huang2024vbench++}. 
Specifically, we computed the correlation between our evaluation results and human annotations. Figure~\ref{fig:win_ratio_dot} and Table~\ref{tab:win_ratio} presents the correlation plot and numerical win ratios respectively, illustrating the alignment between human judgment and VBench-2.0 evaluations in terms of model-level win ratios across video generation models in each dimension. 

\section{Insights and Discussions}
\label{sec:insights}
In this section, we present key insights from VBench-2.0's evaluation, highlighting trade-offs, model characteristics, and in-depth discussion on superficial versus intrinsic faithfulness in video generation.

\subsection{Characteristics of Recent SOTA Models}

From Figure~\ref{fig:fig_paper_radar_big}, we can observe the relative strengths and weaknesses of each model under evaluation.

\smallTitle{Sora-480p~\cite{sora}.} 
Sora clearly excels in \textit{Human Fidelity} and \textit{Creativity} dimensions compared to other SOTA models. It demonstrates a strong ability to generate human figures with reasonable anatomical consistency throughout a video while also showing improvisational skill in producing novel and imaginative content. This makes Sora a potential tool for human-centric filmmaking and artistic exploration. 
However, it falls short in \textit{Controllability}, \textit{Physics} and \textit{Commonsense} dimensions, indicating that the generated videos may not align well with user-provided text prompts and sometimes violate real-world principles.

\smallTitle{Kling 1.6~\cite{kling}.} 
Kling demonstrates relative strengths in the \textit{Commonsense}, \textit{Controllability} and camera-related (\textit{Multi-View Consistency}, \textit{Camera Motion}) dimensions. These capabilities make Kling well-suited not only for tasks that require precise camera control, but also for broader applications involving coherent, accurate, and user-guided visual storytelling or simulation. Additionally, Kling does not show significantly weaker performance in any particular area, suggesting that its training data is broad and well-rounded, making it a valuable reference for future model development.

\smallTitle{CogVideoX-1.5~\cite{yang2024cogvideox}.} 
This model is relatively strong in most dimensions related to complex prompt adherence (\eg \textit{Complex Landscape} and \textit{Complex Plot}) and \textit{Physics}, but shows notably poorer results in human-centric dimensions such as \textit{Human Fidelity} and \textit{Motion Rationality}. These outcomes suggest that CogVideoX’s training data contains limited high-quality human-related content.

\smallTitle{HunyuanVideo~\cite{kong2024hunyuanvideo}.} 
Although HunyuanVideo is relatively weaker in many VBench-2.0 dimensions, it demonstrates impressive strengths in human-related aspects (\textit{Human Fidelity} and \textit{Motion Rationality}). This suggests that HunyuanVideo likely benefits from training data rich in high-quality, human-related content.

\subsection{Key Limitations of Recent SOTA Models}

\smallTitle{Key Challenge: Generating Complex Plots.} 
In the \textit{Complex Plot} dimension, state-of-the-art video generation models struggle to follow detailed text descriptions involving multiple scenes, character interactions, and logical story progression. A major limitation is that current foundation video generative models typically produce single-shot videos under 10 seconds in length, insufficient for conveying coherent narratives. This highlights an important direction for future research: building models capable of functioning more like true filmmakers.

\smallTitle{Surprising Weakness: Controllability in Simple Dynamics.} Most models perform poorly in capturing \textit{Dynamic Spatial Relationships} and \textit{Dynamic Attributes}. Even in relatively simple cases, where an entity’s position, relationship, or attribute (\eg color) is instructed to change, and models fail in about 80\% of the time. These shortcomings, despite the simplicity of the underlying semantics, are likely due to inadequate captioning granularity in video generation datasets. Existing video captioning pipelines may not be intentionally describing how object attributes or positions evolve over time, weakening models' understanding of these dynamics. Enhancing these pipelines with more focused, temporally grounded instructions during video captioning could help address this gap.

\subsection{The Role of Prompt Engineering}
In modern video generation, the Prompt Refiner rewrites or augments input text prompts to enhance video generation quality. For implementation details, please refer to the Sec~\ref{sec:setup}. Below, we highlight several notable and potentially surprising observations.

\noindent\textbf{Controllability vs. Creativity.} Our evaluation results reveal a trade-off between  \textit{Creativity} and \textit{Controllability}. Sora performs well in creative tasks but struggles with controllability, while the other models show the reverse pattern. 
This suggests that models emphasizing creativity and diversity are better at flexibly imagining novel content, but may be less capable of strictly and accurately following control signals.
Alternatively, the Prompt Refiner used with the other three models may improve controllability by fine-graining the text at the expense of diversity. Sora’s internal prompt optimization appears to take a different approach. Going forward, prompt refinement strategies should consider not only precision but also the preservation of diversity, which is essential for creating open-ended content and simulating the distributions of the real world.

\noindent\textbf{Prompting Partially Compensates for Physical Reasoning Gaps.}
Physics is often challenging in video generation. However, with the exception of Sora-480p, models guided by the external Prompt Refiner demonstrated reasonably strong performance in the Physics dimension. This suggests that even without an inherent understanding of physical laws, models can be steered toward physically plausible outcomes through carefully designed prompts. Therefore, the core difficulty may lie less in physical reasoning itself, and more in achieving precise video-text alignment. 

\noindent\textbf{Limited Impact on Knowledge-Driven Dimensions.}
For dimensions that rely on model's intrinsic visual understanding and prior knowledge, like \textit{Human Fidelity}, \textit{Camera Motion}, \textit{Geometry}, and \textit{Commonsense}, we observe no consistent performance trend from models using different Prompt Refiners. These aspects likely extend beyond the scope of logical inference or direct text-to-video mapping, limiting the Prompt Refiner’s influence. This suggests that success in these areas may depend less on prompt engineering and more on underlying data quality and model architecture. 

\subsection{Superficial Faithfulness vs. Intrinsic Faithfulness: Do Not Miss Out on Any Pillar}
\textit{Superficial Faithfulness} (\eg cinematographic quality) often shapes the first impression viewers get from a video. As a result, models that produce aesthetically pleasing and smooth outputs are frequently perceived as ``better''. However, this perception could be misleading. In practice, \textit{Intrinsic Faithfulness}, which includes elements like storytelling, logical progression, and world simulation, is equally important for determining a model's potential for real-world applications. 
For example, as shown in Figure~\ref{fig:fig_paper_radar_big} and Table~\ref{tab:raw_metrics}, CogVideoX performs relatively well across many VBench-2.0 dimensions, though its visual \textit{Quality Score} in VBench suggests room for improvement compared to models like Sora and HunyuanVideo. Conversely, HunyuanVideo produces visually impressive results, though its performance in many structure-driven dimensions in VBench-2.0 suggests opportunities for further growth in those areas.
These observations highlight a common bias toward relying primarily on visual quality when judging video generation models. To address this, we encourage the community to use both VBench and VBench-2.0 together, enabling a more  comprehensive and in-depth evaluation across both \textit{Superficial Faithfulness} and \textit{Intrinsic Faithfulness}.

\subsection{Challenge, Current Solution and Future Work of Video Generation Evaluation}
\noindent\textbf{Challenge.} Since \textit{Intrinsic Faithfulness} is more difficult to accurately evaluate compared to \textit{Superficial Faithfulness} (\ie evaluation models need to have capabilities that include complex scene understanding, commonsense reasoning, and physical world perception), the demand for using large models such as VLMs and LLMs has gradually increased. However, in practical applications, dimensions such as \textit{Human Anatomy} and \textit{Motion Rationality} are areas where current large models cannot serve as reliable evaluators. The main reasons can be summarized into two points: 
1) Training data gap: Real-world video data used for training VLMs does not contain anomaly or unnatural entity states that are commonly present in generated videos;
2) Capability limitations: VLMs perform poorly in aspects such as object counting and 3D understanding.

\noindent\textbf{Current Solution.} In this paper, to address the above two challenges, we construct the corresponding solutions: 1) For different tasks and scenarios, we build anomaly specialists to collect abnormal data regarding human body, motion, entity splitting and merging. This data is then mixed with real datasets for training, achieving high accuracy while maintaining generalizability. 2) Currently, there are two mainstream approaches: fine-tuning-based and non-fine-tuning methods. Fine-tuning-based methods can easily overfit to the generated training data (due to limited data size), achieving high test accuracy. However, since most generated videos are less than 10 seconds in length, models tuned in this way struggle to generalize to longer videos in the future, and the overall generalization ability of the model is also a major concern. Due to the above drawbacks, VBench-2.0 adopts a non-fine-tuning approach. We break down a large problem into multiple visual phenomena that can be understood by VLMs, and design redundant questions to improve the accuracy of VLMs. For dimensions that require high-level understanding ability, we further integrate LLM into the evaluation. 

\noindent\textbf{Future Work.} We believe that some dimensions will inevitably require the understanding capabilities of large models to be resolved in the future. However, other dimensions (particularly those involving content that should not exist in real data) indeed need to be handled by specialized expert models, as outlined in the two points mentioned above. Here, following the above classification, we outline the key directions that future next-generation video generation evaluation methods should explore: 1) Although the types of anomalies judged by different dimensions vary (\textit{Human Anatomy} and \textit{Instance Preservation} evaluate anomalies in human motion, appearance anomalies and entity state anomalies, respectively), there is an overlap between these anomaly types. For example, during the entity splitting process, limb anomalies will also occur. Therefore, integrating abnormal data to train a unified clip-level anomaly detection model can help the model capture both the differences and commonalities across various anomalies, thereby enabling the development of a more generalized and unified anomaly detection large model. 2) Accuracy and sustainability are two important pillar off a good benchmark. Fine-tuning-based methods cannot meet the sustainability requirement under current circumstances (unless transform into a clip-level formulation that is independent of video length and construct a virtual database with the same scale as the VLM training data). Therefore, what needs to be done in the future is to more thoroughly evaluate existing models' understanding of real-world physical laws and 3D scenes (\eg VLM benchmarks for 3D space reasoning~\cite{yang2025thinking}, cinematography understanding~\cite{liu2025shotbench} and physics understanding~\cite{balazadeh2024synthetic, al2024unibench}). When a truly capable and qualified intelligent large model emerges in the future, it will mark the beginning of the next generation of video generation benchmarks.

\section{Conclusion}
While recent video generative models have achieved \textit{superficial faithfulness}, true progress requires advancing towards \textit{intrinsic faithfulness}, ensuring adherence to physical laws, commonsense reasoning, and structured coherence. To address this, we introduced VBench-2.0, a benchmark assessing models on five key dimensions beyond superficial faithfulness. To accurately evaluate these more challenging dimensions, we fully explore the capabilities of state-of-the-art vision-language models (VLMs) and large language models (LLMs). We build generalists in dimensions where they excel, and employ various specialists for assessment in areas where they are less proficient. VBench-2.0 complements VBench by expanding evaluation to deeper aspects of video generation, providing a multi-dimensional and human-aligned evaluation framework. We believe that VBench-2.0 is an important contribution to the video generation community, shaping the field into its next era.

\noindent\textbf{Future Work.} We will continually add more video generative models to VBench-2.0 when they become available.

\noindent\textbf{Potential Negative Societal Impacts.} Although VBench-2.0 does not directly generate videos, the evaluation process inevitably involves working with generated content. As video generative models grow more powerful and capable of producing increasingly realistic scenes, we emphasize the importance of safety and ethical considerations when using these models. We encourage responsible use to mitigate potential risks associated with AI-generated media.

\section*{Acknowledgments}
This study is supported by the Ministry of Education, Singapore, under its MOE AcRF Tier 2 (MOE-T2EP20221-0012, MOE-T2EP20223-0002), and under the RIE2020 Industry Alignment Fund – Industry Collaboration Projects (IAF-ICP) Funding Initiative, as well as cash and in-kind contribution from the industry partner(s).

{
    \small
    \bibliographystyle{IEEEtran}
    \bibliography{main}
}

\begin{IEEEbiography}
[{\includegraphics[width=1in,height=1.25in,clip,keepaspectratio]{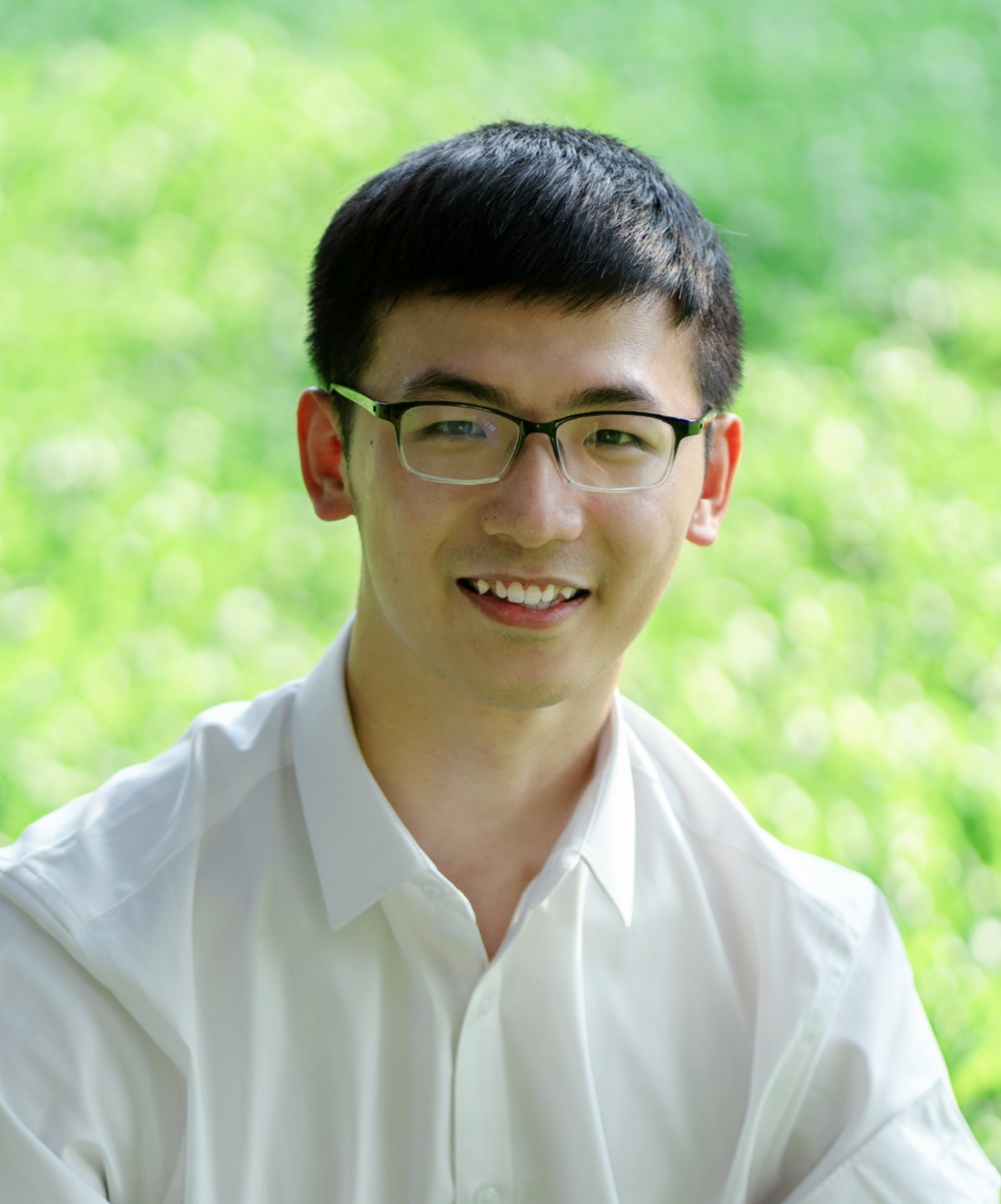}}]{Dian Zheng}  is an incoming Ph.D. student at MMLab@CUHK, Chinese University of Hong Kong (CUHK), supervised by Prof. Hongsheng Li. He received his Master's degree in 2025 and his Bachelor's degree in 2022 from Sun Yat-sen University and Dalian University of Technology respectively. His current research interests focus on visual generation and evaluation.
\end{IEEEbiography}

\begin{IEEEbiography}
[{\includegraphics[width=1in,height=1.25in,clip,keepaspectratio]{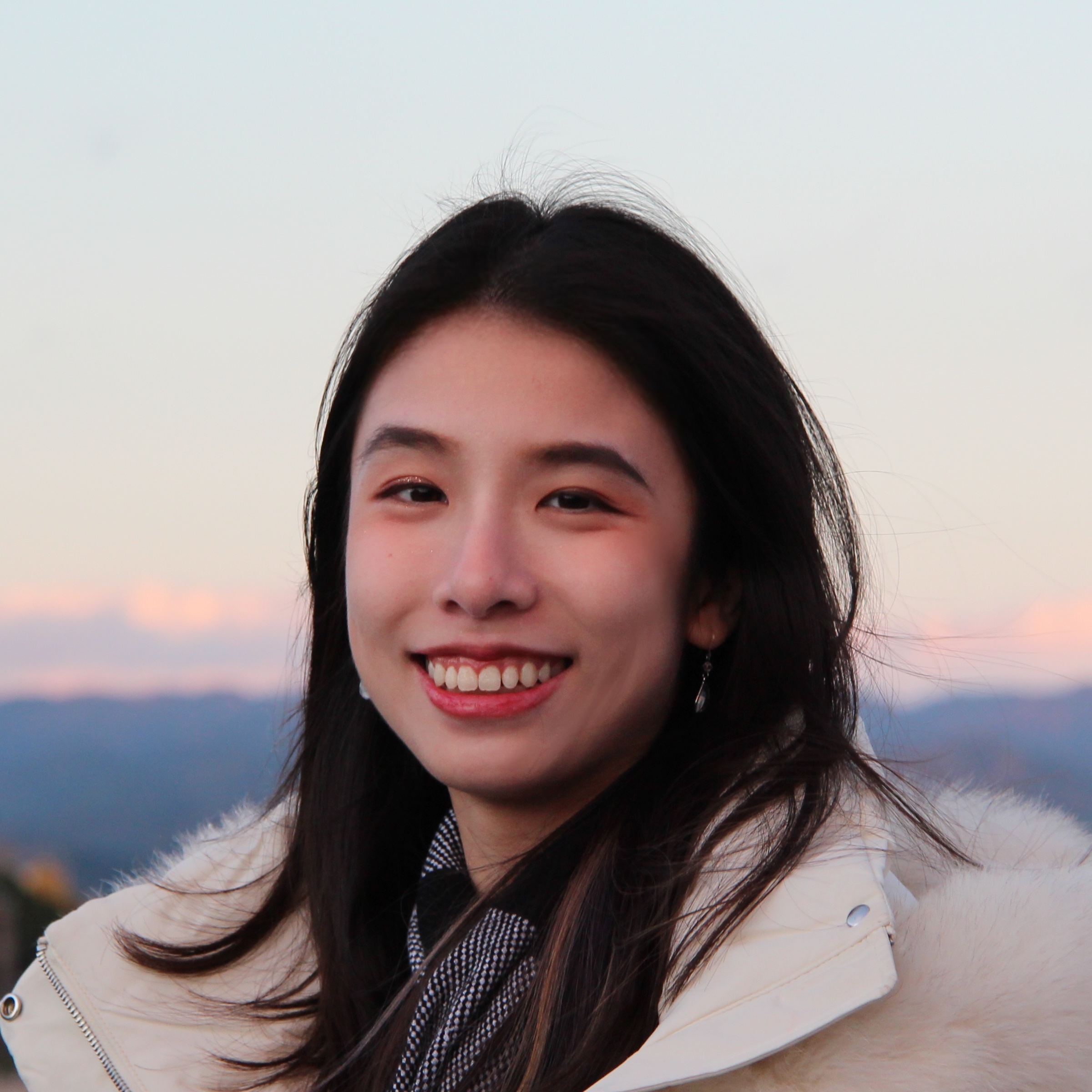}}]{Ziqi Huang} is currently a Ph.D. student at MMLab@NTU, Nanyang Technological University (NTU), supervised by Prof. Ziwei Liu. She received her Bachelor's degree from NTU in 2022. Her current research interests include visual generation and evaluation. She is awarded Google PhD Fellowship 2023, and is a recipient of the 2025 Apple Scholars in AI/ML PhD Fellowship.
\end{IEEEbiography}

\begin{IEEEbiography}
[{\includegraphics[width=1in,height=1.25in,clip,keepaspectratio]{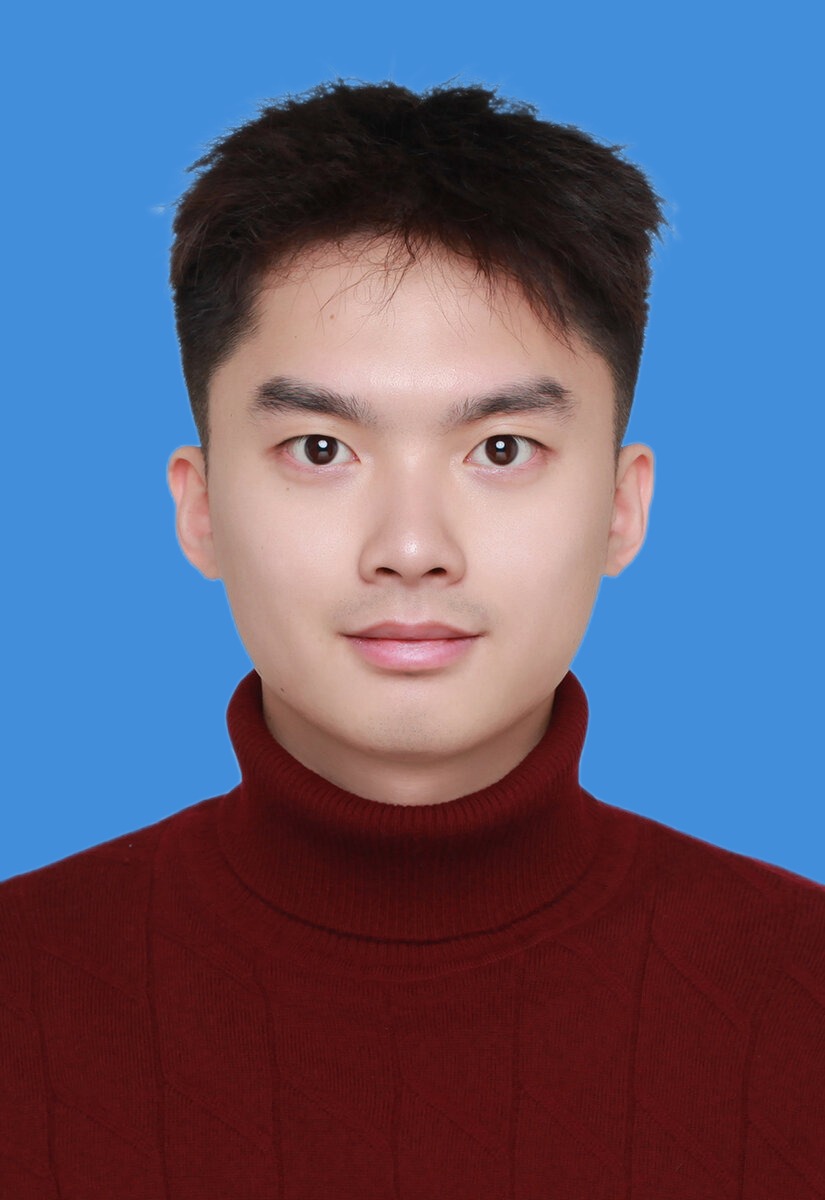}}]
{Hongbo Liu} is currently pursuing his Ph.D. degree at Tongji University, under the supervision of Prof. Shengjie Zhao. He received his Bachelor's degree from Tongji University in 2023. His research interests lie in multimodal understanding and generation.
\end{IEEEbiography}

\begin{IEEEbiography}
[{\includegraphics[width=1in,height=1.25in,clip,keepaspectratio]{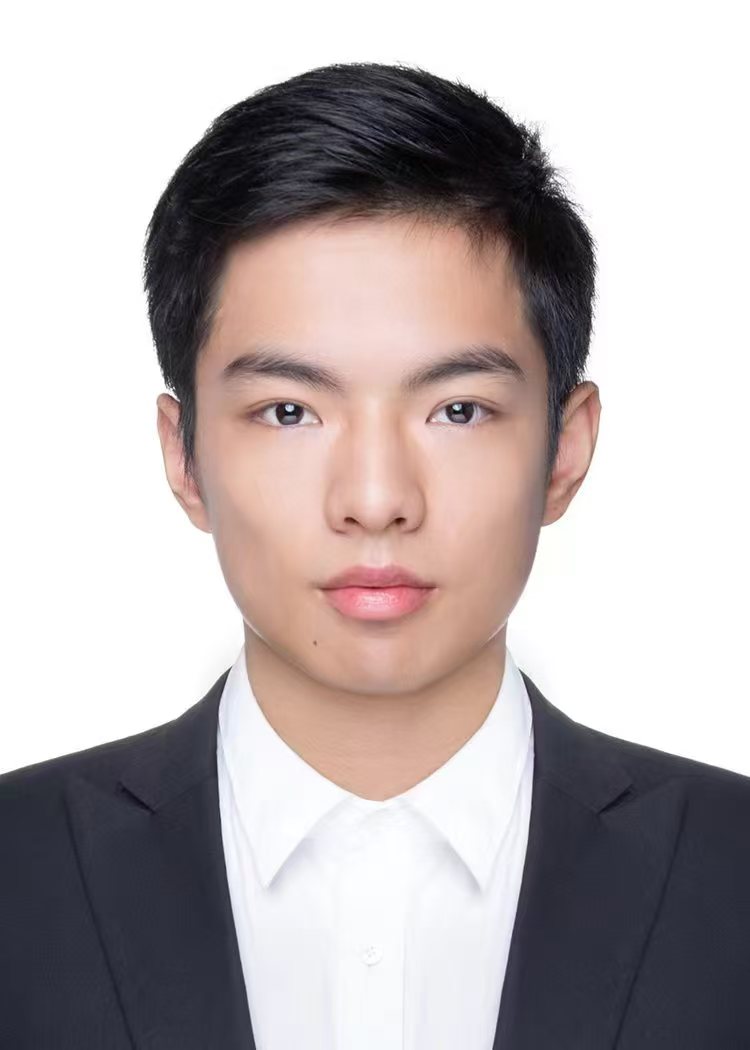}}]
{Kai Zou} is currently a Master's student at the University of Science and Technology of China, under the supervision of Professor Bin Liu. He obtained his Bachelor's degree from Shanghai University in 2023. His current research interests lie in multimodal generation and understanding.
\end{IEEEbiography}

\begin{IEEEbiography}
[{\includegraphics[width=1in,height=1.25in,clip,keepaspectratio]{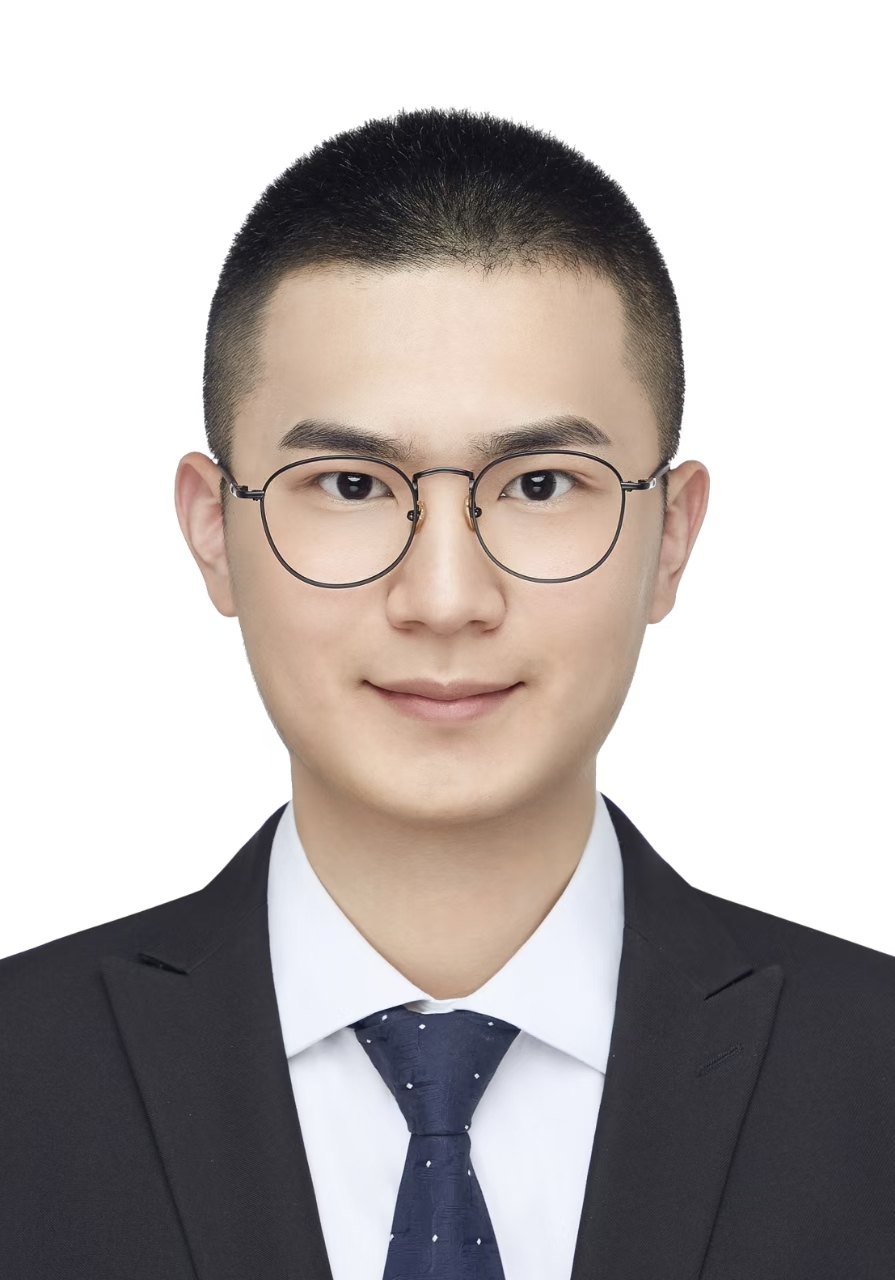}}]
{Yinan He} is currently a Research Engineer at Shanghai AI Laboratory, where he is a member of the OpenGVLab. He received his Master’s degree from Beijing University of Posts and Telecommunications. His current research interests include video understanding and multi-modality large language models.
\end{IEEEbiography}

\begin{IEEEbiography}
[{\includegraphics[width=1in,height=1.25in,clip,keepaspectratio]{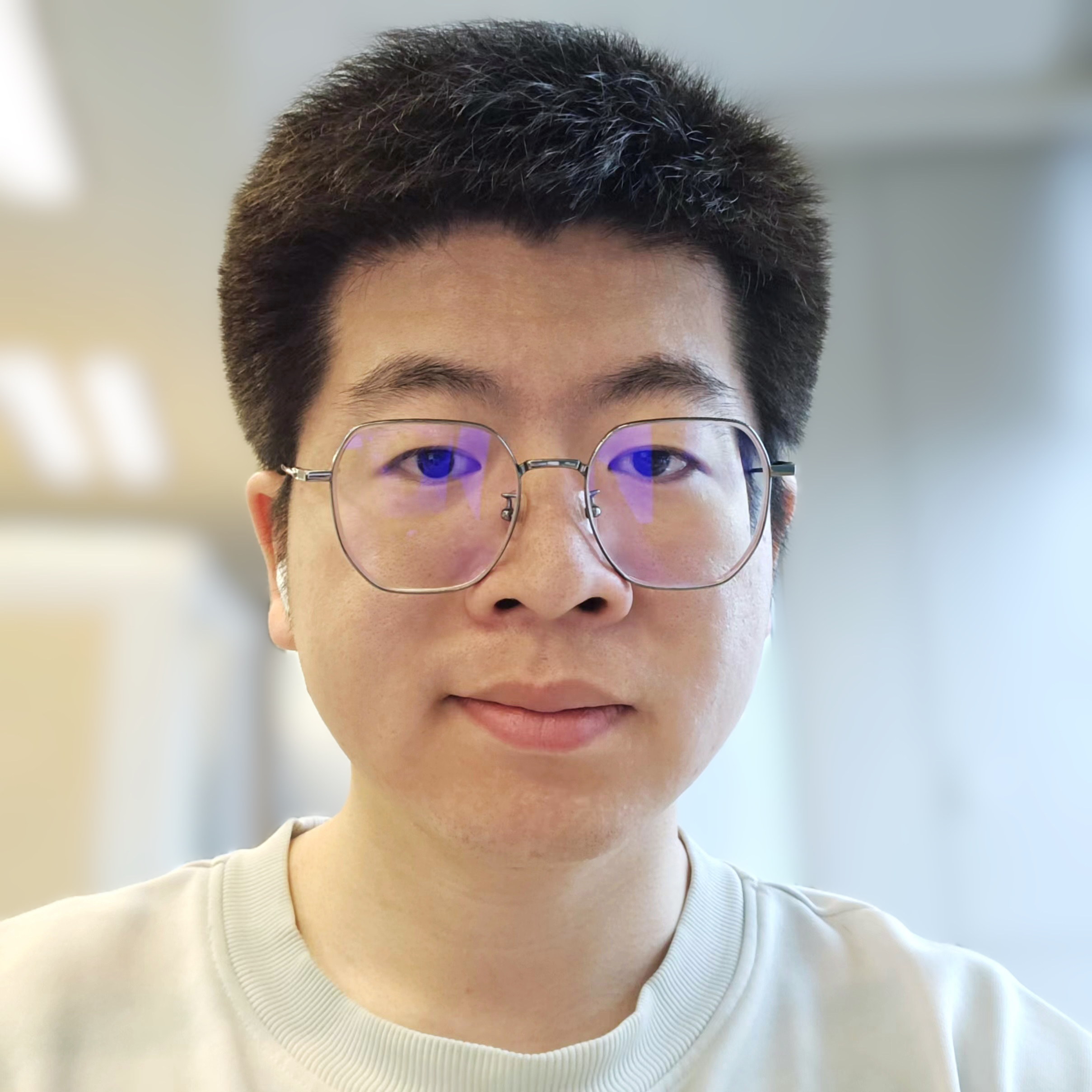}}]{Fan Zhang} is currently a research assistant at the Shanghai Artificial Intelligence Laboratory, supervised by Prof. Ziwei Liu. He received his bachelor's degree from the University of Electronic Science and Technology of China (UESTC). His research interests include generative models and computer vision.
\end{IEEEbiography}

\begin{IEEEbiography}
[{\includegraphics[width=1in,height=1.25in,clip,keepaspectratio]{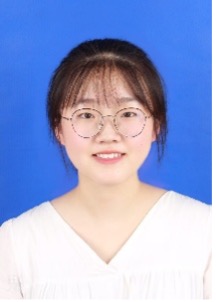}}]{Lulu Gu} is a research intern at the Shanghai Artificial Intelligence Laboratory. She is currently pursuing her master’s degree at the School of Electronic Information and Electrical Engineering, Shanghai Jiao Tong University. Her research interests lie in deep learning and generative modeling.
\end{IEEEbiography}

\begin{IEEEbiography}
[{\includegraphics[width=1in,height=1.25in,clip,keepaspectratio]{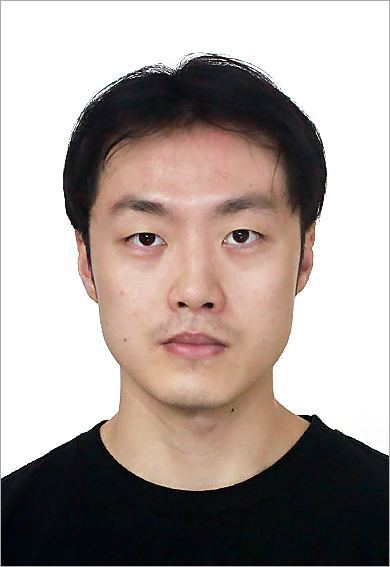}}]
{Yuanhan Zhang} is currently a Ph.D. student at MMLab@NTU, Nanyang Technological University, supervised by Prof. Ziwei Liu. His interests lie in computer vision and deep learning. In particular, He is focused on adapting foundation models---from vision to multi-modal---for real-world exploration. He has published several papers in ICCV, ECCV, CVPR, NeurIPS and {\em IEEE
 Transactions on Pattern Analysis and Machine Intelligence} (TPAMI). He also served as a reviewer for CVPR, ICCV, ECCV, NeurIPS, ICML, ICLR, {\em IEEE
 Transactions on Pattern Analysis and Machine Intelligence} (TPAMI), and {\em International Journal of Computer Vision} (IJCV).
\end{IEEEbiography}

\begin{IEEEbiography}
[{\includegraphics[width=1in,height=1.25in,clip,keepaspectratio]{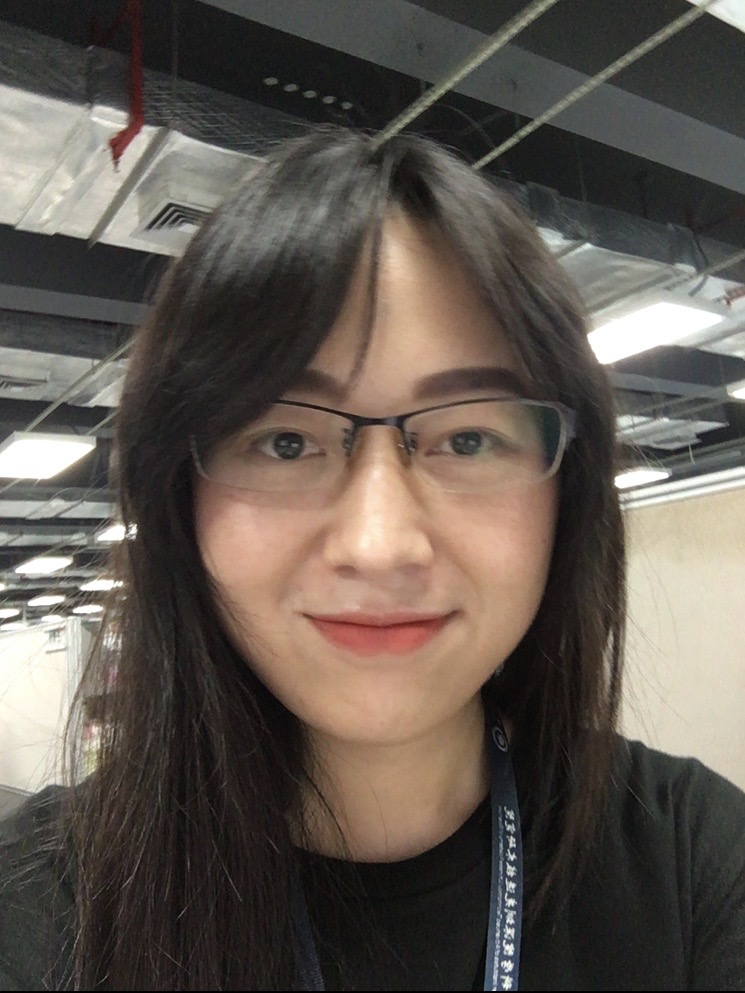}}]
{Jingwen He} is currently a Ph.D. student at Chinese University of Hong Kong, supervised by Prof. Wanli Ouyang. She received her M.Phil. degree in electronic
and information engineering from the University of Sydney, Australia, in 2019, supervised by Prof. Dong Xu. Her research interests include image generation, video generation, and vision language models.
\end{IEEEbiography}

\begin{IEEEbiography}
[{\includegraphics[width=1in,height=1.25in,clip,keepaspectratio]{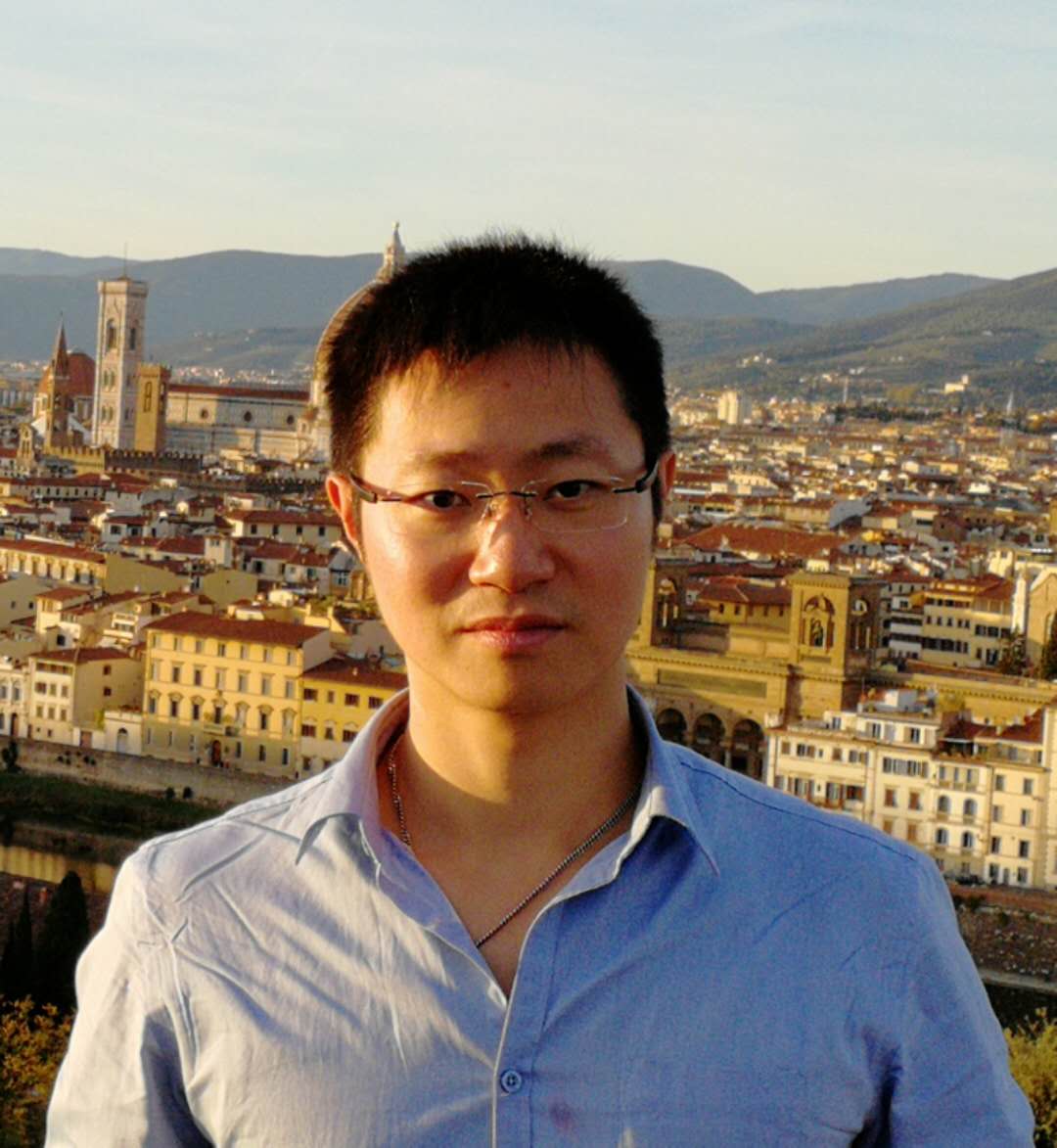}}]{Wei-Shi Zheng} is now a full Professor with Sun Yat-sen University. His research interests include person association and activity understanding, AI robotics learning, and the related weakly supervised/unsupervised and continuous learning machine learning algorithms. He has now published more than 200 papers, including more than 150 publications in main journals (TPAMI, IJCV, SIGGRAPH, TIP) and top conferences (ICCV, CVPR, ECCV, NeurIPS). He has ever served as area chairs of ICCV, CVPR, ECCV, BMVC, NeurIPS and etc. He is associate editors/on the editorial board of IEEE-TPAMI, Artificial Intelligence Journal, Pattern Recognition. He has ever joined Microsoft Research Asia Young Faculty Visiting Programme. He is a Cheung Kong Scholar Distinguished Professor, a recipient of the NSFC Excellent Young Scientists Fund, a Fellow of IAPR and a recipient of the Royal Society-Newton Advanced Fellowship of the United Kingdom.
\end{IEEEbiography}

\begin{IEEEbiography}
[{\includegraphics[width=1in,height=1.25in,clip,keepaspectratio]{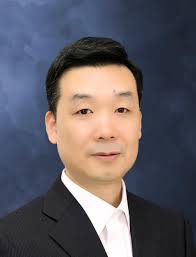}}]{Yu Qiao} (Senior Member, IEEE) is a professor and Leading Scientist at the Shanghai AI Laboratory, previously the Director of the Institute of Advanced Computing and Digital Engineering at the Shenzhen Institutes of Advanced Technology, Chinese Academy of Science. He has published more than 300 research papers with more than 100,000 citations. His team won the AAAI 2021 Best Paper, the CVPR 2023 Best Paper, and the ACL 2024 Distinguished Paper. He received the Young Scholar Award of Wang Xuan Award, the First Prize of the Guangdong Technological Invention Award, and the Jiaxi Lv Young Researcher Award from the Chinese Academy of Sciences. His research interests include deep learning on computer vision, video understanding and generation, and multimodal large models.
\end{IEEEbiography}

\begin{IEEEbiography}
[{\includegraphics[width=1in,height=1.25in,clip,keepaspectratio]{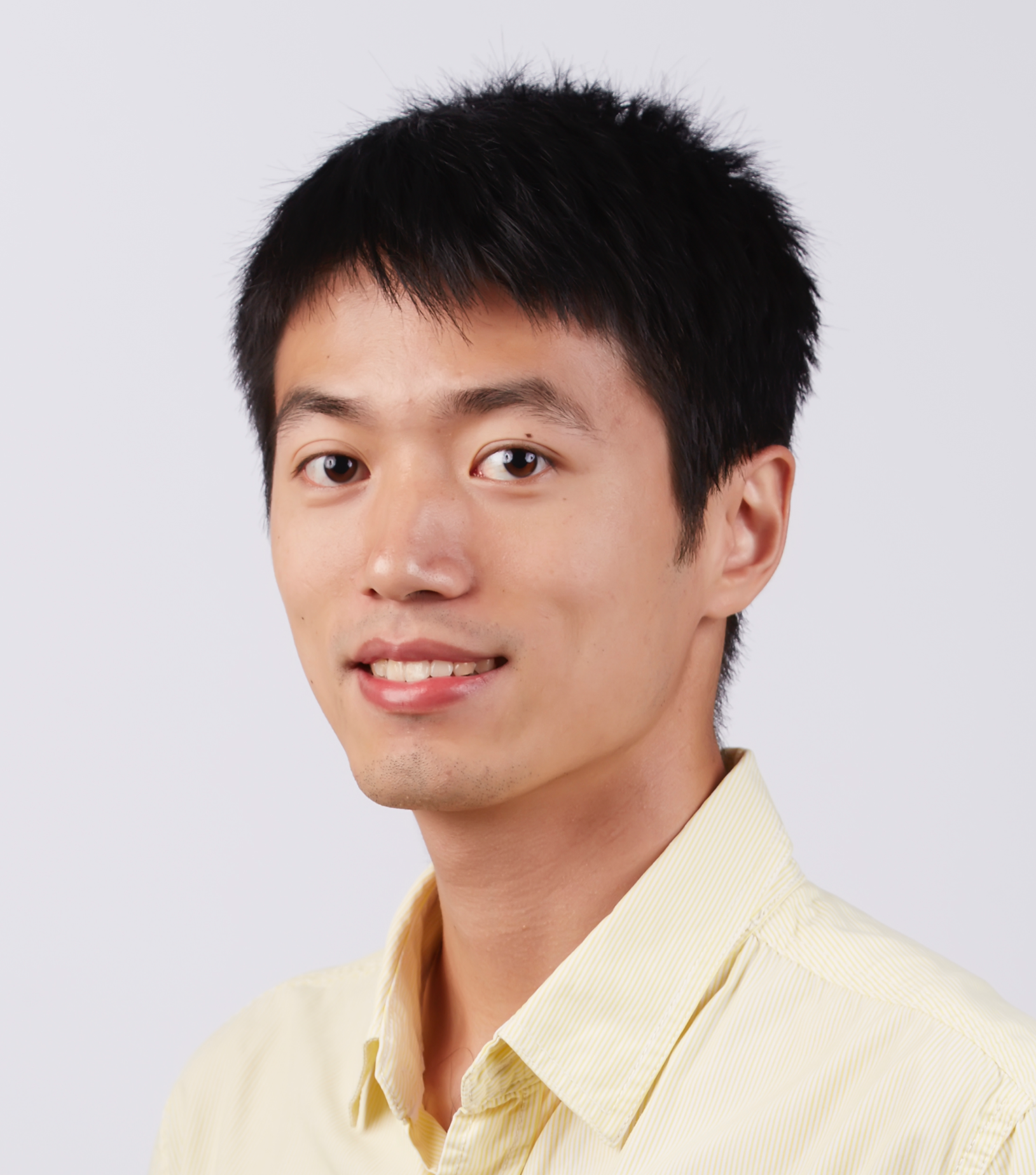}}]{Ziwei Liu} is currently an Associate Professor at Nanyang Technological University, Singapore. His research revolves around computer vision, machine learning and computer graphics. He has published extensively on top-tier conferences and journals in relevant fields, including CVPR, ICCV, ECCV, NeurIPS, ICLR, SIGGRAPH, TPAMI, TOG and Nature - Machine Intelligence. He is the recipient of PAMI Mark Everingham Prize, CVPR Best Paper Award Candidate, Asian Young Scientist Fellowship, International Congress of Basic Science Frontiers of Science Award and MIT Technology Review Innovators under 35 Asia Pacific. He serves as an Area Chair of CVPR, ICCV, ECCV, NeurIPS and ICLR, as well as an Associate Editor of IJCV.
\end{IEEEbiography}

\end{document}